\pdfoutput=1
\documentclass[11pt]{article}
\usepackage[preprint]{acl}
\usepackage{times}
\usepackage{latexsym}
\usepackage[T1]{fontenc}
\usepackage[utf8]{inputenc}
\usepackage{microtype}
\usepackage{inconsolata}
\usepackage{amsmath,amssymb}
\usepackage{graphicx}
\usepackage{needspace,placeins}
\usepackage{stfloats}
\usepackage{booktabs,array}
\usepackage{capt-of}
\usepackage{xcolor}
\usepackage{url}

\newcommand{\rev}[1]{#1}
\newcommand{\revnew}[1]{#1}

\title{Conceptizer: Next-Token Prediction Is Not All You Need}
\author{%
  Haoran Zheng\\
  The University of Chicago\\
  \texttt{haoranzheng@uchicago.edu}
  \And
  Chen Shani\\
  Stanford University\\
  Tel Aviv University\\
  \texttt{cshani@tauex.tau.ac.il}
}
\newcommand{\ConceptBuildFigureVersionFive}{%
\begin{figure*}[t]\centering
\includegraphics[width=\linewidth]{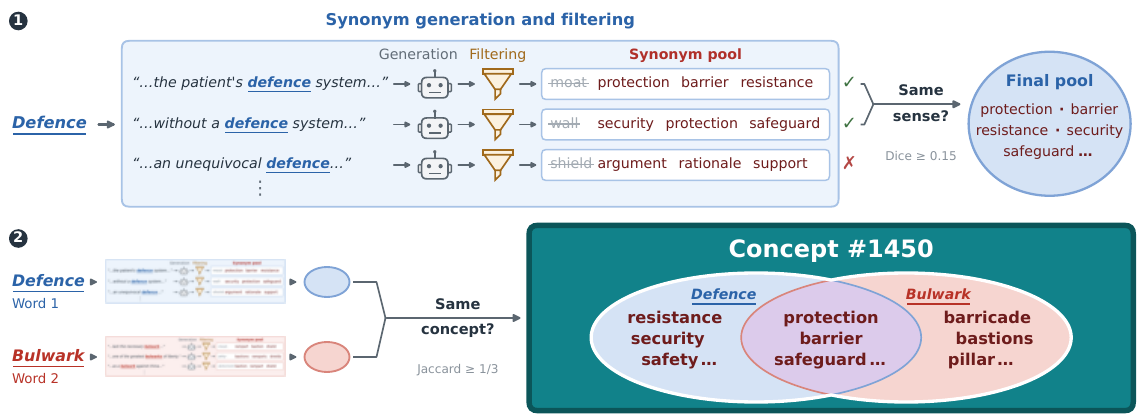}
\caption{\textbf{Building Conceptizer's fixed, model-agnostic concept vocabulary.} For up to 25 FineWeb-Edu occurrences of each content word, LLMs and WordNet propose synonym candidates, and an LLM judge keeps those that fit the sentence. Dice similarity groups one word's synonym pools into senses, and Jaccard similarity merges the senses of different words into 9,061 concepts (Section~\ref{sec:data}; Appendices~\ref{app:data} and~\ref{app:sensitivity}).}\label{fig:concept-build}
\end{figure*}}
\newcommand{\ConceptTrainFigureVersionFive}{%
\begin{figure*}[t]\centering
\includegraphics[width=\linewidth]{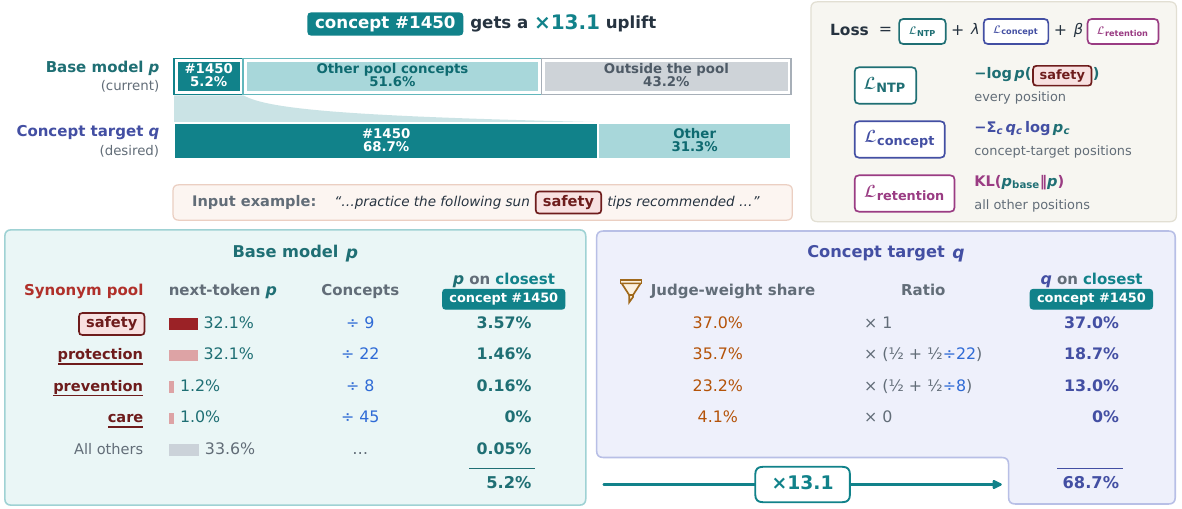}
\caption{\textbf{Concept training with Conceptizer.} Concepts: each token's probability is split evenly over the concepts it has a membership in. Ratio $\kappa_j$: the fraction of each word's judge-weight share $\rho_j$ sent to the closest concept $c^\star$, \#1450. Uplift: $q \div p$ on \#1450. \emph{Care} is not in \#1450, so its $p$ and $q$ on \#1450 are 0; all its concepts are other pool concepts. Other pool concepts: all other concepts that \emph{protection}, \emph{prevention} and \emph{care} belong to; outside the pool: the remaining concepts, and tokens in no concept, such as \emph{-related}.}\label{fig:concept-train}
\end{figure*}}
\newcommand{\WordExampleFigure}{%
\begin{figure}[t]\centering
\includegraphics[width=\linewidth]{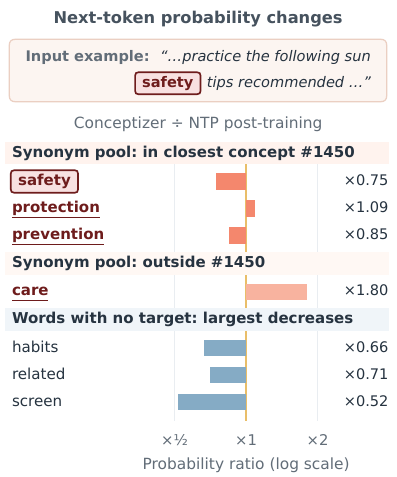}
\caption{\textbf{Conceptizer raises the next-token probability of \emph{protection} and \emph{care} and lowers that of \emph{safety} and \emph{prevention}.} After 50,000 steps (Llama 1B, seed 0).}\label{fig:word-example}
\end{figure}}
\newcommand{\PerplexityMatrixFigure}{%
\begin{figure*}[t]\centering
\includegraphics[width=\linewidth]{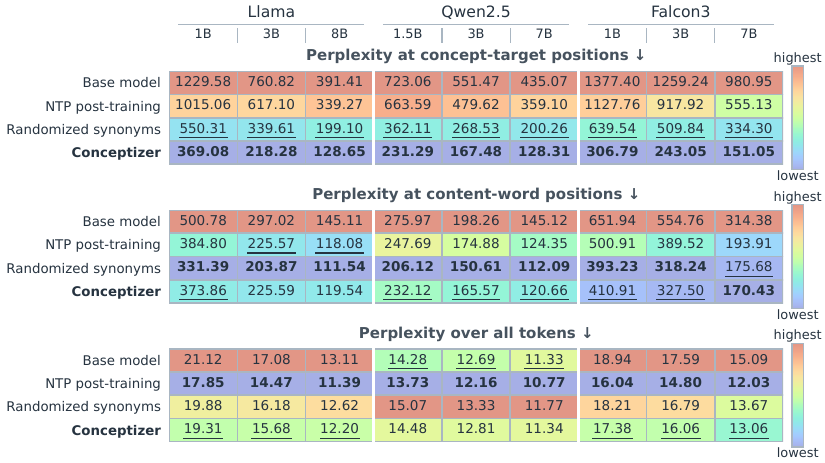}
\caption{\textbf{Conceptizer has the lowest perplexity against the 3 baselines at concept-target positions in every model.} NTP post-training has the lowest global perplexity in every model, and Randomized synonyms the lowest on content words in 8 of 9. Held-out means of training seeds 0, 37 and 42. Colors scale per model and panel.}\label{fig:ppl-matrix}
\end{figure*}}
\newcommand{\DownstreamFigure}{%
\begin{figure*}[t]\centering
\includegraphics[width=\linewidth]{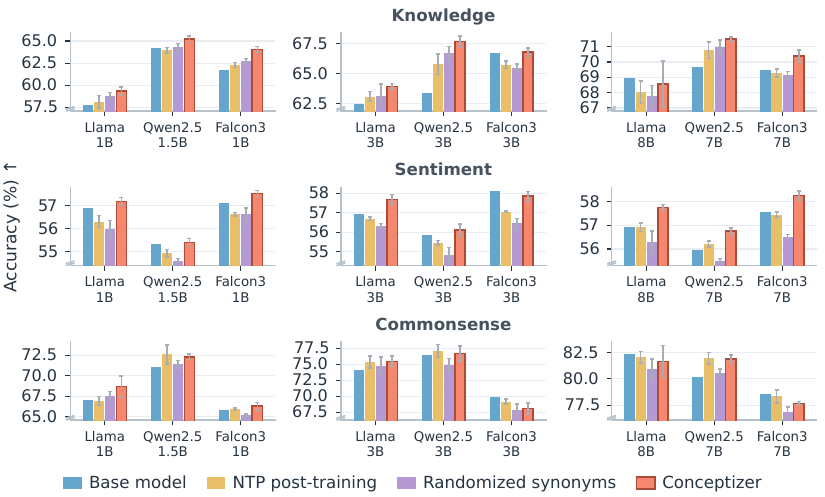}
\caption{\textbf{Conceptizer leads on knowledge and sentiment tasks in 8 of 9 models each.} Commonsense is comparable. Means of seeds 0, 37 and 42; error bars: each method's 95\% Student-$t$ training-seed interval.}\label{fig:downstream}
\end{figure*}}
\newcommand{\MtebFigure}{%
\begin{figure*}[tp]\centering
\includegraphics[width=\linewidth]{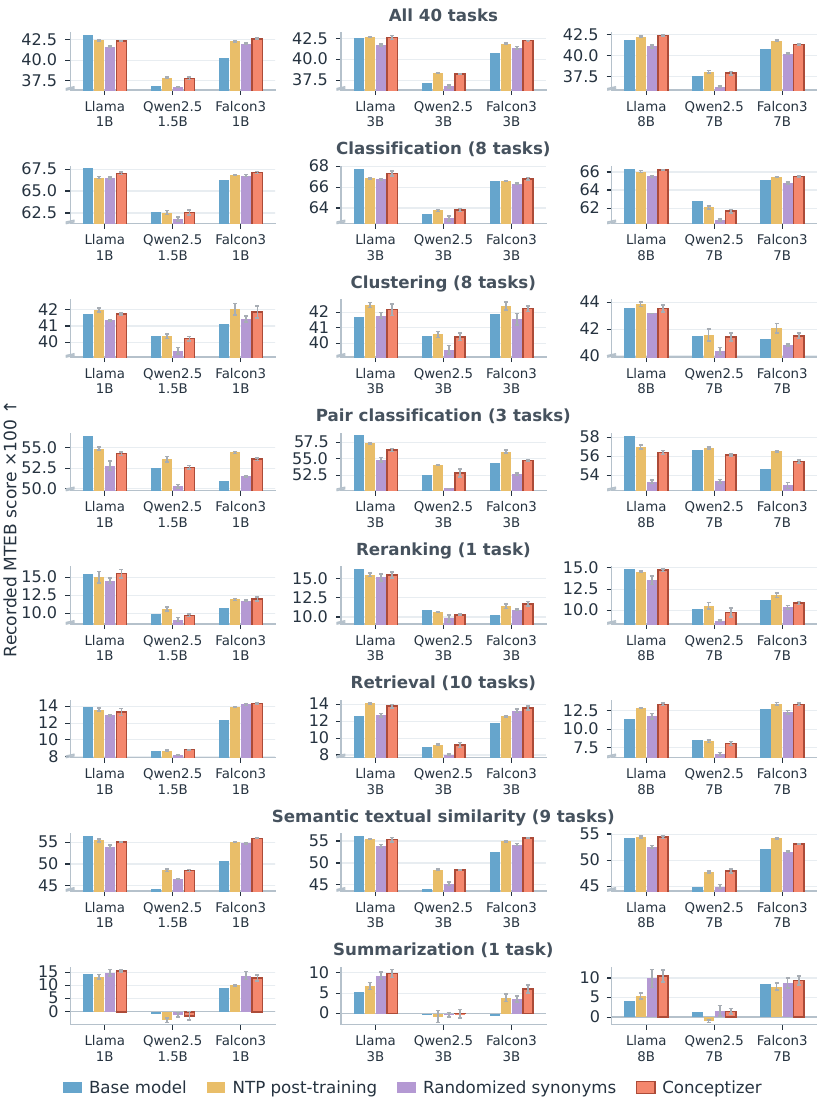}
\caption{\textbf{By MTEB section, no method has the best mean throughout.} Means of seeds 0, 37 and 42; error bars: each method's 95\% Student-$t$ training-seed interval (values in Table~\ref{tab:mteb}).}\label{fig:mteb}
\end{figure*}}
\newcommand{\SizeAccuracyFigure}{%
\begin{figure*}[tp]\centering
\includegraphics[width=\linewidth]{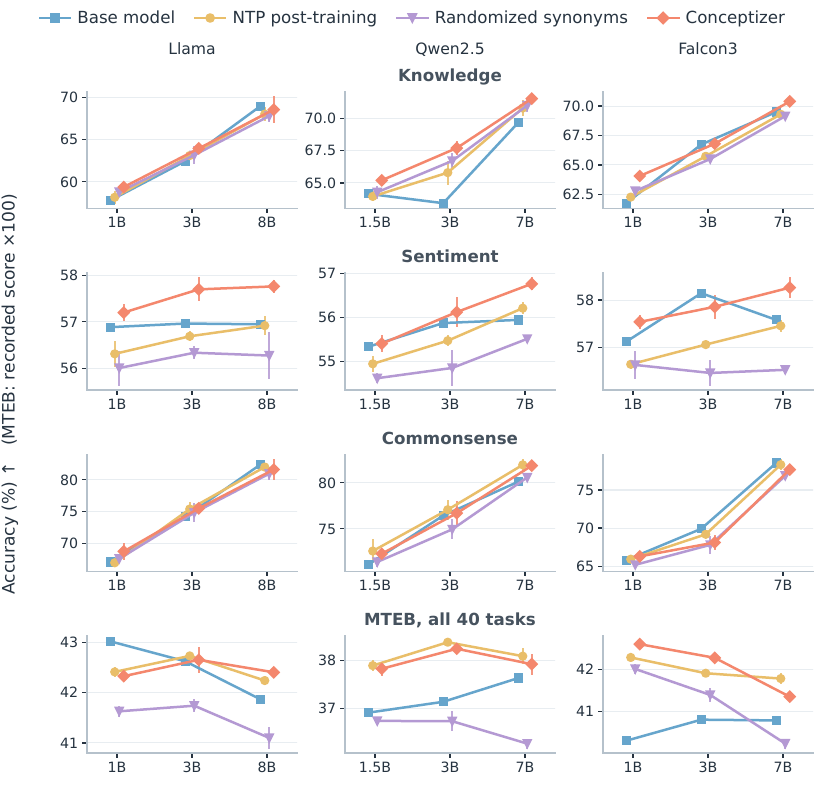}
\caption{\textbf{Downstream performance against model size, by family.} Means of training seeds 0, 37 and 42 with 95\% Student-$t$ intervals (Tables~\ref{tab:seed-knowledge}, \ref{tab:seed-sentiment}, \ref{tab:seed-commonsense} and~\ref{tab:mteb}).}\label{fig:size-accuracy}
\end{figure*}}
\newcommand{\SizeSynonymFigure}{%
\begin{figure}[t]\centering
\includegraphics[width=\linewidth]{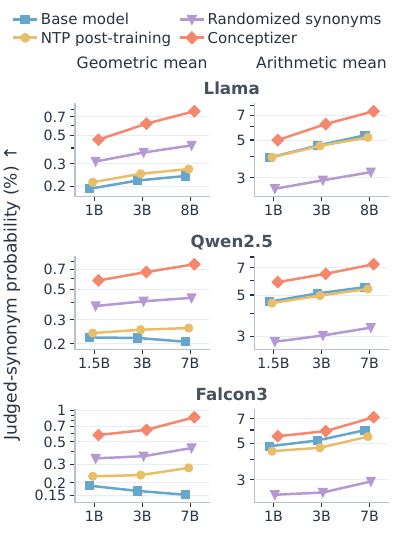}
\caption{\textbf{Conceptizer has the highest judged-synonym probability in every model, by both means, and its probability rises with model size in every family.} Seed 0, means over held-out concept-target positions (Table~\ref{tab:synonym-probability}); logarithmic axes.}\label{fig:size-synonyms}
\end{figure}}
\newcommand{\ConstructionSweepFigure}{%
\begin{figure*}[t]\centering
\includegraphics[width=\linewidth]{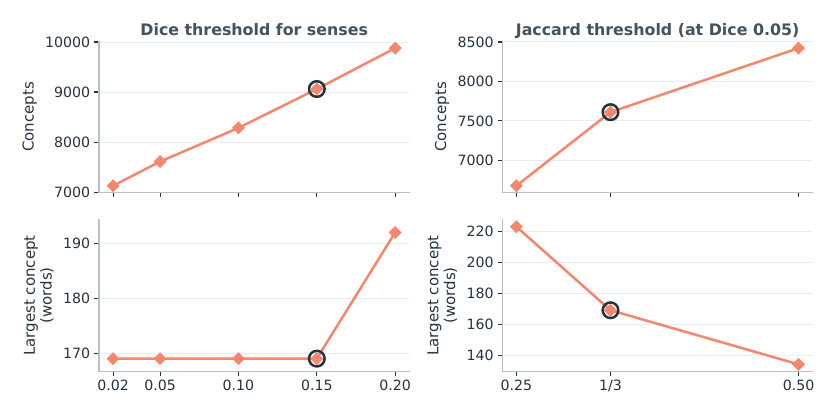}
\caption{\textbf{A higher Dice threshold gives more concepts, and a higher Jaccard threshold gives more and smaller concepts.} The counts of Table~\ref{tab:sensitivity} (top), one setting changed at a time; ringed: the adopted value.}\label{fig:construction-sweep}
\end{figure*}}
\newcommand{\TrainingCurvesFigure}{%
\begin{figure*}[tp]\centering
\includegraphics[width=\linewidth]{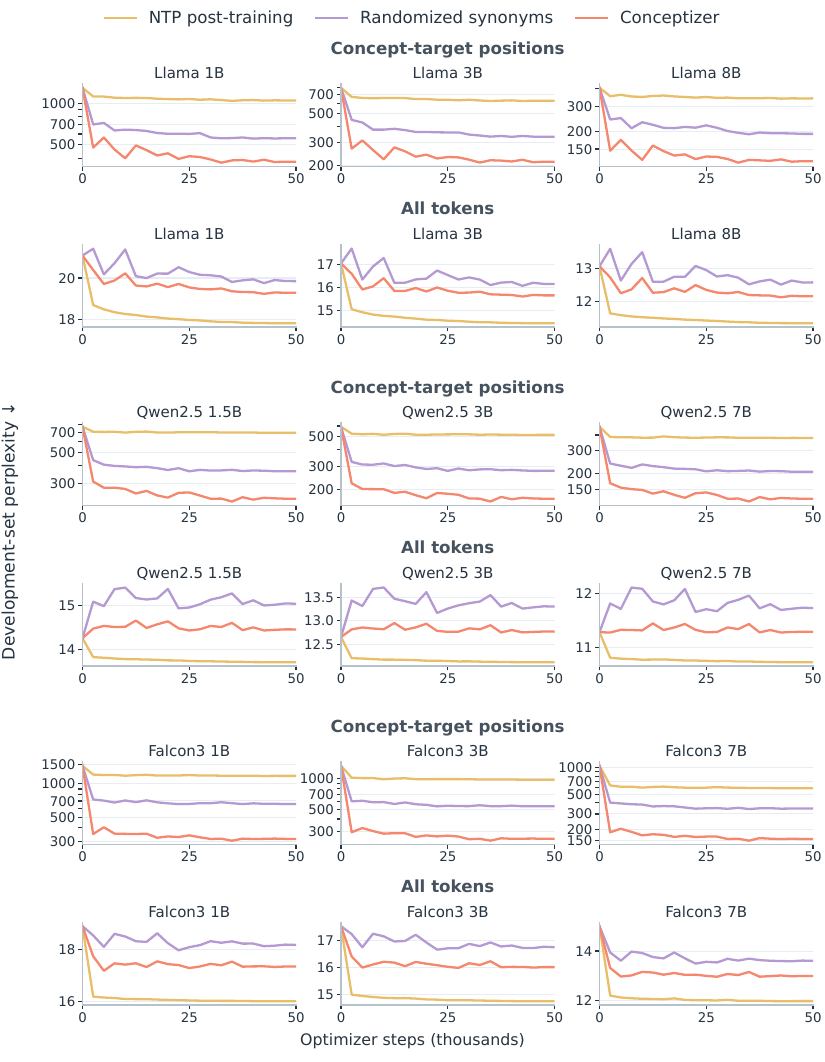}
\caption{\textbf{Development-set perplexity during training.} Scored every 2,500 optimizer steps; step 0 is the base model. Lines: means of training seeds 0, 37 and 42; concept-target rows on logarithmic axes.}\label{fig:training-curves}
\end{figure*}}

\begin{document}
\maketitle
\begin{abstract}Next-token prediction (NTP) rewards \textbf{only} the token that occurs next, penalizing every alternative. But natural language often offers many ways to express the same meaning. We introduce Conceptizer, a training framework that extends next-token prediction with a reward for \textit{meaning}. Conceptizer uses a fixed, model-agnostic vocabulary of \rev{$\sim$9K} concepts, built from contextual synonyms, as units of meaning alongside tokens as units of text. It maps each next-token distribution to a concept distribution and shares the reward for the target meaning across its possible wordings. This provides a concept-level training signal without adding model parameters. Applied to \rev{3} model families at 1B-8B parameters, Conceptizer achieves the lowest held-out NTP perplexity at concept-target positions and assigns the highest probability to contextual synonyms\rev{, against all 3 baselines}. It also improves downstream performance on \rev{knowledge} and \rev{sentiment} tasks while remaining comparable on \rev{commonsense}.
 These results show that language models can be directly optimized to predict meaning, not just the next token.
\end{abstract}
\section{Introduction}Next-token prediction (NTP) directs a language model (LM) toward a single target at each position: the observed next token in the corpus. After ``…practice the following sun'', a base model gives \emph{protection} and \emph{safety} 32\% probability each, yet NTP rewards only the observed word, \emph{safety}, and pushes \emph{protection} down as it would a wrong word.

The units that NTP scores are built from surface statistics, not meaning: standard tokenizers partition text based on statistical regularities rather than semantic relationships. We introduce \emph{Conceptizer},\footnote{Code, the concept vocabulary and the training data: \url{https://github.com/HaoranZhengRaul/conceptizer-code}.} a training framework that adds a fixed vocabulary of units of meaning alongside each model's existing tokenizer. In this vocabulary, \emph{safety}, \emph{protection} and \emph{security} can share concept \#1450, allowing the training objective to reward different wordings of the same contextual meaning.

Offline, we construct a vocabulary of 9,061 concepts (Fig.~\ref{fig:concept-build}). Unlike prior work, this vocabulary is \rev{built once from \emph{contextual synonyms} (lexical substitutes; \citealp{mccarthy2007lexsub})} and shared across models. \rev{During training, a \revnew{\emph{concept loss}} turns each next-token distribution into a distribution over concepts and pulls it toward a target built from the next token's contextual synonyms (Fig.~\ref{fig:concept-train}).}

Inference is unchanged from standard NTP: the final softmax and \rev{token} vocabulary remain the same. The only difference is that during training, \textbf{Conceptizer \rev{rewards} all tokens that express the same underlying concept}.

We post-train \rev{Llama 3}, Qwen2.5 and Falcon3 models \citep{meta2024llama32,grattafiori2024llama3,qwen2024qwen25,falcon2024falcon3} at 3 scales (1B to 8B).
\rev{We then compare Conceptizer with 3 baselines: the base model, next-token prediction post-training (NTP post-training) and a Randomized synonyms control.}
At held-out concept-target positions, Conceptizer assigns higher NTP probability to the observed word and its contextual synonyms than all \rev{3} baselines \rev{(Fig.~\ref{fig:ppl-matrix}; Appendix~\ref{app:controls})}.
On downstream evaluations, Conceptizer outperforms the baselines in \rev{26 of 27} comparisons on \rev{knowledge} and \rev{sentiment} tasks, while remaining competitive on \rev{commonsense} \rev{(Fig.~\ref{fig:downstream})}.

\section{Related Work}\paragraph{\rev{Concept-aware language models.}}
\rev{\citet{shani2023conceptaware} group a model's predictions into concepts, \citet{iyer2026beyond} post-train a model to predict a noun's synonyms and hypernyms, and \citet{zhang2026concepts} extend this to all content words, pooling the probability of contextual synonyms drawn from the model's own predictions, one occurrence at a time. Conceptizer instead builds a fixed concept vocabulary shared by every word, occurrence and model, and its concept loss contains the losses of \citeauthor{iyer2026beyond} and \citeauthor{zhang2026concepts} as special cases (Appendix~\ref{app:math}).}

\paragraph{\rev{Tokenization.}}
\rev{Token vocabularies are built from surface statistics, not meaning: BPE merges frequent pairs of symbols \citep{gage1994bpe,sennrich2016bpe}, WordPiece and unigram tokenization choose units by likelihood \citep{schuster2012wordpiece,wu2016gnmt,kudo2018subword}, and SentencePiece and byte-level BPE apply them to raw text and bytes \citep{kudo2018sentencepiece,radford2019gpt2}. So no token vocabulary connects \emph{safety}, \emph{protection} and \emph{security}. Conceptizer keeps each model's tokenizer and adds concepts alongside it.}

\paragraph{Words grouped by meaning.}
Surface forms of one answer compete for probability \citep{holtzman2021surface}, and semantic entropy pools the answers that share a meaning to estimate a model's uncertainty \citep{kuhn2023semantic}. Grouping words by meaning has a long lineage: class-based language models factor a word's probability through its class \citep{brown1992class,goodman2001classes}; SenseBERT also predicts one of WordNet's 45 supersenses through a learned head, pooling the probability of the supersenses WordNet allows for the word, independent of its context \citep{levine2020sensebert,miller1995wordnet}; LIBERT adds the classification of word relations such as synonymy to BERT's pretraining \citep{lauscher2020libert}; and Backpack models learn sense vectors per word inside the network \citep{hewitt2023backpack}. Conceptizer finds each word's senses from its lexical substitutes \citep{mccarthy2007lexsub} across sentences, as in substitute-based word sense induction \citep{amrami2018wsi,eyal2022wsi}, and merges the senses of different words into concepts. A model then reaches these concepts through concept memberships, without new parameters, and each target comes from its \rev{occurrence's} own contextual synonyms.

\paragraph{Soft targets.}
Softening the one-hot target also has a long history: label smoothing mixes it with a uniform distribution \citep{szegedy2016inception}, semantic variants spread the credit over related tokens or sequences \citep{elbayad2018smoothing,lukasik2020semantic}, D2GPo adds a KL term toward a Gaussian prior over tokens ordered by embedding similarity \citep{li2020d2gpo}, and AdaLabel learns a context-dependent token target \citep{wang2021adalabel}. Distillation trains toward a teacher's distribution \citep{hinton2015distilling,kim2016distillation}, as our retention loss does toward the frozen base model, and selective language modeling keeps the loss only on the tokens a reference model marks as useful \citep{lin2024rho1}. These losses score each token or sequence on its own, while the concept loss scores the total probability of each concept. Partial-label learning also scores a set of candidate labels together, but only one of them is correct \citep{cour2011partial,cid2012partial}.

\paragraph{Concepts as units of generation.}
Other work predicts or reasons with ``concepts'' and other latent units, each under its own definition of a concept. Large Concept Models take a sentence's embedding as a concept and predict the next one \citep{lcm2024large}. CoCoMix takes features of a sparse autoencoder as concepts, predicts them and mixes them into the model's hidden states \citep{tack2026cocomix}. ConceptLM takes a learned code for each chunk of 4 tokens as a concept and predicts it to guide the next tokens \citep{liu2026conceptlm}. Soft Thinking, Coconut and Soft Tokens, Hard Truths reason by feeding back a continuous vector, a probability-weighted mixture of token embeddings or the model's last hidden state, in place of the token sampled at each step \citep{zhang2025softthinking_neurips,hao2025coconut,butt2026softtokens}. These ``concepts'' are learned or continuous representations, not lists of words. For Conceptizer, a concept is an explicit set of words that can replace one another in a context without changing its meaning, such as \#1450's \emph{safety}, \emph{protection} and \emph{security}.

\ConceptBuildFigureVersionFive %
\ConceptTrainFigureVersionFive
\section{Data Preparation: Concept Vocabulary and Targets}\label{sec:data}

\paragraph{Data.} We use FineWeb-Edu, a collection of English educational web documents \citep{penedo2024fineweb}. We tag nouns, verbs and adjectives with spaCy \citep{honnibal2020spacy} and refer to these as \emph{content words}. For each word, we collect up to 25 occurrences with the same part of speech (POS), grouping all morphological forms together: \emph{documented} and \emph{documenting}, for example, are both treated as (document; verb). We discard (word; POS) combinations with fewer than 3 occurrences. This yields 2.33 million occurrences of 172,925 words (13.5 per word on average). The concept vocabulary $\mathcal C$ is constructed from 88,367 of these occurrences. Let $w_0$ denote the \emph{observed word}, such as \emph{safety} after ``…practice the following sun'' \rev{(our running example)}.

\paragraph{Synonym candidate generation.}
\rev{For each collected occurrence,} \rev{3} sources propose \emph{synonym candidates}\rev{. Llama-3.3-70B-Instruct \citep{meta2024llama33} proposes contextual replacements for the sense in the sentence, while} Qwen2.5-72B-Instruct \citep{qwen2024qwen25} and WordNet \citep{miller1995wordnet} see only the word and its POS \rev{and cover all its senses. This casts a wide net, which the LLM judge narrows.}

\paragraph{LLM judge.}
All candidates are input into an \emph{LLM judge}, a frozen gpt-oss-120B \citep{openai2025gptoss}. It is instructed to keep the candidates that can replace \rev{the observed word} $w_0$ without changing the sentence's meaning or grammatical form.
\rev{The judge's answers vary between runs (temperature 0.7), so one run can miss a borderline synonym. We run it 3 times and keep any candidate one run keeps} \rev{(details in Appendix~\ref{app:data})}.
\rev{In the same answer, the judge can add up to 2 contextual synonyms to avoid obvious misses by the candidate sources.}
These kept words form \rev{the occurrence's} \emph{synonym pool} (``concept set'' in \citealp{zhang2026concepts}).
\rev{Of 2.33 million occurrences, 1.84 million have a synonym pool; the remainder are mostly names, codes and specialized terms such as \emph{insulin}, for which no suitable candidate replacement is found.}
For simplicity, synonyms that the trained model's tokenizer splits into several tokens are dropped, and the rest, at most 10 \citep[the cap of][]{zhang2026concepts}, are labeled $w_1,\dots,w_m$.
For example, for (safety; noun) \rev{after ``…practice the following sun''}, the judge keeps 3 of 43 candidates: \{$w_1=$~\emph{protection}, $w_2=$~\emph{prevention}, $w_3=$~\emph{care}\} (it \rev{cut} \emph{security}, among others).

\paragraph{Judge weights.}
\rev{Each synonym candidate is then weighted by how often the LLM judge keeps it when it is proposed, across all collected occurrences of $w_0$ (up to 25). The 3 judge runs of any single occurrence give each candidate one binary verdict, not a fraction such as 2/3.} If a synonym $w_j$ is proposed in $n_j$ of these occurrences and kept in $k_j$, its \emph{judge weight} $r_j$ is the Laplace-smoothed rate
\begin{equation}
 r_j=\frac{k_j+1}{n_j+2},
 \label{eq:judge-weight}
\end{equation}
and $w_0$ receives $r_0=1$. Over the 25 occurrences of \emph{safety} (Table~\ref{tab:safety_judgments} in Appendix~\ref{app:data}), \emph{protection} is proposed and kept in all 25 ($r_1=26/27=0.96$), \emph{prevention} is proposed in 6, once by the judge, and kept in 4 ($r_2=5/8=0.63$), and \emph{care} is kept in 2 of 25 ($r_3=0.11$). Smoothing keeps thin evidence from looking certain: \emph{carefulness}, kept the one time it was proposed, gets a judge weight $r_j=2/3$ instead of 1. A synonym's judge weight $r_j$ is the same for all collected occurrences of $w_0$, but each occurrence uses it only for its own synonym pool: \emph{security}, kept in \rev{22 of 25}, goes unused for ``…practice the following sun'', where the judge cut it.

\paragraph{The concept vocabulary.}
\rev{We build the \revnew{\emph{concept vocabulary}} $\mathcal C$ from the occurrences of 5,939 frequent content words (3.8\% of all collected; Fig.~\ref{fig:concept-build}; Appendix~\ref{app:data-concepts}). First, to distinguish their senses, we group each word's synonym pools into \emph{senses} by Dice similarity ($2|A\cap B|/(|A|+|B|)\ge0.15$ for pools $A$ and $B$; \citealp{dice1945measures}). A sense's \emph{final pool} is \rev{the word} itself plus the union of its qualified synonym pools. Second, to create each \emph{concept}, we merge senses of different words by Jaccard similarity ($|A\cap B|/|A\cup B|\ge1/3$ for final pools $A$ and $B$, with at least 2 shared words; \citealp{jaccard1912distribution}). A concept's \revnew{\emph{member words}} are the union of these final pools.} The resulting concept vocabulary \rev{$\mathcal C=\{c_0,\dots,c_{9060}\}$} ($c_n$ is written \#$n$) contains \rev{49,387} member words, \rev{far exceeding the original 5,939,} and is model-agnostic: the same 9,061 concepts serve all 9 models we post-train.

\paragraph{Concept memberships.}
Concept memberships connect each model's output tokens to our shared concept vocabulary \rev{$\mathcal C$}. The vocabulary is the same across models, but their tokenizers differ, so the same word may correspond to different tokens in different models. We therefore construct the mapping separately for each tokenizer.

A token belongs to a concept if the corresponding word appears in that concept. We also include a token when it is the first token of a longer word in \rev{that concept (a \emph{first-token membership})}. A token can belong to multiple concepts. For example, the token for \emph{safety} belongs to \rev{9 concepts, among them \#1450 and \revnew{\#6387, which} lists \emph{safetylines} but not \emph{safety}}. In total, Falcon3 has \rev{23,992} words represented as single tokens that participate in at least one concept, compared with \rev{14,836} for Llama and \rev{14,829} for Qwen2.5.

Let $\mathcal V$ be the model's \rev{output token vocabulary} and $\mathcal H(v)$ be \rev{the \emph{concept memberships} of token $v$, the set of concepts it is associated with}. \rev{A token in no concept has $\mathcal H(v)=\{\bot\}$, the \emph{no-concept outcome},} giving $\mathcal C^+=\mathcal C\cup\{\bot\}$. When a token belongs to multiple concepts, \rev{its \emph{membership weights} $\omega_{vc}$ divide} its contribution equally among them:
\begin{equation}
\omega_{vc}=\frac{\mathbf 1[c\in\mathcal H(v)]}{|\mathcal H(v)|},
\qquad v\in\mathcal V,\ c\in\mathcal C^+.
\label{eq:membership}
\end{equation}

\paragraph{The concept target.}
At each selected \emph{concept-target position} \rev{$i$}, Conceptizer adds a \rev{\emph{concept target} $\mathbf q_i$,} a target distribution over \rev{$\mathcal C^+$ (the concept vocabulary $\mathcal C$ and $\bot$)}, to the standard next-token target. These positions contain content words that are represented as a single whole-word token and have an associated concept (Appendix~\ref{app:data-eligibility}).

We construct $\mathbf q_i$ from the observed word $w_0$ and its judge-weighted contextual synonyms $w_1,\dots,w_m$. First, we identify the \emph{closest concept} $c^\star$: among the concepts containing $w_0$, it is the one that contains \rev{the most contextual synonyms $w_1,\dots,w_m$}. Each word $w_j$ contributes its normalized judge weight \rev{(\emph{judge-weight share}) $\rho_j=r_j/\sum_{l=0}^{m}r_l$,} where $r_j$ is its judge weight (Equation~\ref{eq:judge-weight}).

\rev{We then distribute each word's share $\rho_j$ across its associated concepts $\mathcal H(w_j)$. The observed word $w_0$ assigns all of its share $\rho_0$ to $c^\star$ \rev{($\sigma_0=1$)}. A synonym whose token also belongs to $c^\star$ assigns half of its share $\rho_j$ to $c^\star$ \rev{($\sigma_j=1/2$)} and distributes the remaining half equally across all its concept memberships $\mathcal H(w_j)$, $c^\star$ included. A synonym whose token does not belong to $c^\star$ \rev{($\sigma_j=0$)} distributes its entire share $\rho_j$ equally across its concept memberships $\mathcal H(w_j)$. Thus, the target combines direct evidence for the closest concept $c^\star$ with the other meanings associated with each synonym}:
\begin{equation}
\mathbf q_i=\sum_{j=0}^{m}\rho_j\big[\sigma_j\,\mathbf e_{c^\star}+(1-\sigma_j)\,\boldsymbol\omega_{w_j}\big],
\label{eq:target}
\end{equation}
where $\mathbf e_{c^\star}$ is \rev{$c^\star$'s unit vector}, $\boldsymbol\omega_{w_j}=(\omega_{w_jc})_{c\in\mathcal C^+}$ gives the membership weights of $w_j$'s token (Equation~\ref{eq:membership}), and \rev{$\sigma_j$ is the part of $\rho_j$ sent directly to $c^\star$}.

The resulting $\mathbf q_i$ is a probability distribution over \rev{$\mathcal C^+$ ($\mathcal C$ and $\bot$)}, sums to 1 and is fixed by the training data rather than the model's predictions. Depending on the tokenizer, 298,227 to 475,558 training positions qualify for a concept target. The concept loss is applied at these positions.

\section{Training Conceptizer}\label{sec:method}
We post-train a base model with a LoRA adapter \citep{hu2022lora} on FineWeb-Edu text, using 3 losses: NTP $\mathcal L_{\mathrm{NTP}}$ at every position, a concept loss $\mathcal L_{\mathrm{concept}}$ at concept-target positions $\mathcal S$ and a \revnew{\emph{retention loss}} $\mathcal L_{\mathrm{retention}}$ at all other positions $\mathcal U$\rev{, which keeps the model's next-token distribution there close to the base model's}.

\paragraph{Model concept distribution.}
At prediction position $i$, the logits $\mathbf z_i$ give \rev{the} next-token distribution $\mathbf p_{\mathrm w,i}=\operatorname{softmax}(\mathbf z_i)$ over the \rev{output token vocabulary} $\mathcal V$. The membership weights $\omega_{vc}$ of Equation~\ref{eq:membership} split each token's probability evenly over its concepts $\mathcal H(v)$, and summing per concept gives the \emph{model concept distribution} $\mathbf p_{\mathrm c,i}$ that sums to 1:
\begin{equation}
 p_{\mathrm c,i}(c)=\sum_{v\in\mathcal V}\omega_{vc}\,p_{\mathrm w,i}(v),
 \qquad c\in\mathcal C^+.
 \label{eq:projection}
\end{equation}

\paragraph{Losses and objective.}
In one optimizer step, let $\mathcal I$ be the positions that predict a next token, $g_i$ the observed next token at position $i$ ($g_i=w_0$ at a concept-target position), $\mathcal S\subseteq\mathcal I$ the concept-target positions, $\mathcal U=\mathcal I\setminus\mathcal S$ the other positions, and $\mathbf p^{\mathrm{base}}_{\mathrm w,i}$ the frozen base model's next-token distribution. NTP, $\mathcal L_{\mathrm{NTP}}$, is the cross-entropy of the observed next token $g_i$, the concept loss $\mathcal L_{\mathrm{concept}}$ the cross-entropy between the concept target $\mathbf q_i$ (Equation~\ref{eq:target}) and the model concept distribution $\mathbf p_{\mathrm c,i}$ (Equation~\ref{eq:projection}), and the retention loss $\mathcal L_{\mathrm{retention}}$ the KL divergence between the base model's and the model's next-token distributions, $\mathbf p^{\mathrm{base}}_{\mathrm w,i}$ and $\mathbf p_{\mathrm w,i}$:
\begin{align}
 \mathcal L_{\mathrm{NTP}}&=-\frac{1}{|\mathcal I|}\sum_{i\in\mathcal I}\log p_{\mathrm w,i}(g_i),
 \label{eq:ntp}\\
 \mathcal L_{\mathrm{concept}}&=-\frac{1}{|\mathcal S|}\sum_{i\in\mathcal S}\sum_{c\in\mathcal C^+}q_i(c)\log p_{\mathrm c,i}(c),
 \label{eq:concept-loss}\\
 \mathcal L_{\mathrm{retention}}&=\frac{1}{|\mathcal U|}\sum_{i\in\mathcal U}D_{\mathrm{KL}}\big(\mathbf p^{\mathrm{base}}_{\mathrm w,i}\,\Vert\,\mathbf p_{\mathrm w,i}\big),
 \label{eq:retention}
\end{align}
and the objective $\mathcal L$ is
\begin{equation}
 \boxed{\mathcal L=\mathcal L_{\mathrm{NTP}}+\lambda\,\mathcal L_{\mathrm{concept}}+\beta\,\mathcal L_{\mathrm{retention}}},
 \label{eq:loss}
\end{equation}
with loss weights $\lambda=0.6$ and $\beta=0.5$, held constant across models and methods (hyperparameter sweeps in Appendix~\ref{app:sensitivity}). Each loss is averaged over its own positions in the optimizer step: each concept target $\mathbf q_i$ enters the objective $\mathcal L$ with weight $\lambda/|\mathcal S|$ and each NTP term with $1/|\mathcal I|$.

\paragraph{How the concept loss acts.}
At the running example (Fig.~\ref{fig:concept-train}), the judge weights $r_0,\dots,r_3$ of \emph{safety}, \emph{protection}, \emph{prevention} and \emph{care}, 1, 0.96, 0.63 and 0.11, sum to 2.70, giving judge-weight shares $\rho_0,\dots,\rho_3$ of 37\%, 36\%, 23\% and 4\%. The concept target puts $q_i(c_{1450})=68.7\%$ on the closest concept $c^\star$, \#1450: all of \emph{safety}'s share, half of \emph{protection}'s and \emph{prevention}'s, and 1/22 and 1/8 of their other halves. The remaining 31.3\% lies on the 68 other concepts of \emph{protection}, \emph{prevention} and \emph{care}. At the first optimizer step, the model still equals the base model, because its LoRA adapter starts at zero. The model concept distribution then gives \#1450 only $p_{\mathrm c,i}(c_{1450})=5.23\%$, mostly from \emph{safety}'s and \emph{protection}'s 32.1\% each, split over 9 and 22 concepts. At that step, the gradient of the concept loss $\mathcal L_{\mathrm{concept}}$ acts in 3 ways.

First, like any cross-entropy (Equation~\ref{eq:concept-loss}), the concept loss $\mathcal L_{\mathrm{concept}}$ pulls each concept $c$ by its \textbf{uplift}, \rev{the concept target divided by the model concept distribution,} $q_i(c)/p_{\mathrm c,i}(c)$. For \#1450 the uplift is $68.7/5.23=13.1$: the target $\mathbf q_i$ asks 13.1 times what the model concept distribution $\mathbf p_{\mathrm c,i}$ gives. A concept is \emph{under-served} when the model concept distribution $\mathbf p_{\mathrm c,i}$ gives it less than the target $\mathbf q_i$ asks (uplift above 1) and \emph{over-served} when it gives more (uplift below 1; Appendix~\ref{app:walkthrough}). A concept outside the target $\mathbf q_i$ has an uplift of 0.

Second, the concept memberships $\mathcal H(v)$ pass each concept's uplift down to its tokens. A token $v$ takes from each of its concepts an \emph{uplift share}, the concept's uplift divided by the token's number of concepts $|\mathcal H(v)|$, because the token gives each of its concepts that fraction of its probability in the model concept distribution $\mathbf p_{\mathrm c,i}$ (Equation~\ref{eq:projection}). The sum of the token $v$'s uplift shares over all its concepts $\mathcal H(v)$ is its \textbf{pull}, $\sum_{c}\omega_{vc}\,q_i(c)/p_{\mathrm c,i}(c)$. \emph{Safety}'s token, in 9 concepts, takes an uplift share of $13.1/9=1.46$ from \#1450, and its pull is 1.58. \emph{Protection}'s token, in 22 concepts, takes only $13.1/22=0.60$ from \#1450, and its pull is 1.06. A token with few concepts is pulled strongly, and one with many barely. The token's \textbf{push}, the change the loss asks for in its logit $z_i(v)$, is $p_{\mathrm w,i}(v)\,(\text{pull}-1)$, where the $-1$ comes from the softmax: the logit $z_i(v)$ rises when $v$'s pull is above 1 and falls when \rev{it} is below 1. \emph{Safety}'s push is $0.321\times(1.58-1)=+0.19$, and \emph{protection}'s $0.321\times(1.06-1)=+0.02$.

Third, the concept loss $\mathcal L_{\mathrm{concept}}$ rewards or penalizes the words of a concept together. Every token of \#1450 takes an uplift share from \revnew{it}, not only \revnew{those} in this \rev{occurrence's} synonym pool. \emph{Security}, cut by the judge in this \rev{occurrence}, has a token in 9 concepts, \#1450 among them, so it takes the uplift share of 1.46 and \revnew{its logit} rises. \emph{Security}'s push is only $+0.0004$, against \emph{safety}'s $+0.19$, because the base model gives it only 0.07\%. Next-token prediction and the concept losses of \citet{iyer2026beyond} and \citet{zhang2026concepts} are special cases of the Conceptizer concept loss $\mathcal L_{\mathrm{concept}}$ (Appendix~\ref{app:math}).

\WordExampleFigure
Pulls and pushes change at every step as the model's probabilities move, so the first step does not predict the end of training. \rev{Read} at the same training position \rev{after 50,000 steps} (Fig.~\ref{fig:word-example}; details in Appendix~\ref{app:configuration}), Llama 1B's Conceptizer adapter gives \emph{protection} 42.11\% against \rev{NTP post-training}'s 38.55\%, \emph{care} 1.85\% against 1.03\% and \emph{safety} 25.54\% against 34.02\%. Together, \emph{safety} and its contextual synonyms $w_1,w_2,w_3$ hold 71.35\% of the next-token probability \rev{$\mathbf p_{\mathrm w,i}$} after training.

\section{Implementation Setup}\label{sec:experiments}
\PerplexityMatrixFigure
\paragraph{Models.}
We post-train 9 models: Llama-3.2-1B/3B-Instruct \citep{meta2024llama32} and Llama-3.1-8B-Instruct \citep{grattafiori2024llama3}, Qwen2.5-1.5B/3B/7B-Instruct \citep{qwen2024qwen25}, and Falcon3-1B/3B/7B-Instruct \citep{falcon2024falcon3}.

\paragraph{Methods.}
The \textbf{Base model} is the released checkpoint, \textbf{\rev{NTP post-training}} \rev{trains it} with NTP and retention losses, $\mathcal L_{\mathrm{NTP}}$ and $\mathcal L_{\mathrm{retention}}$, and \textbf{Conceptizer} adds the concept loss $\mathcal L_{\mathrm{concept}}$. \textbf{Randomized synonyms} is post-trained like Conceptizer, but each concept-target position gets the contextual synonyms $w_1,\dots,w_m$ and judge weights $r_j$ of another target with as many synonyms. The position's closest concept $c^\star$ is also redrawn uniformly from the observed token's concepts $\mathcal H(w_0)$, so the observed word $w_0$'s share \rev{$\rho_0$} goes to a random one.

\paragraph{Training.}
\rev{All post-trained methods share the same training text, order and recipe: 50,000 optimizer steps of a LoRA adapter on FineWeb-Edu, with seeds 0, 37 and 42} (Appendix~\ref{app:configuration}). \rev{Concept-target positions} are 0.15 to 0.24\% of all positions.

\section{Experiments \texorpdfstring{\rev{and}}{and} Results}\label{sec:results}
\DownstreamFigure

\subsection{Prediction on Held-Out Text}
\label{sec:heldout}

\rev{We hold out 10\% of the documents from training (Appendix~\ref{app:data}) and split them between development and final evaluation. NTP perplexity is measured on the final part, 15,909 to 15,920 documents and 9.74 to 9.81 million predicted tokens per family, at 3 sets of positions:} concept-target positions (5,154 for Llama and Qwen2.5, 10,028 for Falcon3), \revnew{\emph{content-word positions}} (each occurrence's first token) and all tokens (\revnew{\emph{global perplexity}}).

\paragraph{Results.}
\textbf{At held-out concept-target positions, Conceptizer gives the observed word $w_0$ the most probability against the 3 baselines in every model} (Fig.~\ref{fig:ppl-matrix}; Table~\ref{tab:seed-perplexity}): in geometric mean, \revnew{2.6 to 3.8} times \rev{NTP post-training}'s, 1.5 to 2.2 times Randomized synonyms' and 3.0 to 6.5 times the base model's. For Llama 1B, the observed word $w_0$'s geometric-mean probability over the 5,154 held-out concept-target positions rises \revnew{2.8} times, from 0.10\% under \rev{NTP post-training} to 0.27\% (perplexity, its inverse, from 1,015 to 369).

\textbf{Conceptizer also gives the contextual synonyms $w_1,\dots,w_m$ the most probability against the 3 baselines in every model} (seed 0; Appendix~\ref{app:controls}). Conceptizer's \emph{judged-synonym probability}, the total probability of a position's contextual synonyms $w_1,\dots,w_m$ without the observed word $w_0$, is in geometric mean \revnew{2.2 to 3.1} times \rev{NTP post-training}'s, 1.5 to 2.0 times Randomized synonyms' and 2.4 to 5.6 times the base model's (for Llama 1B, 0.464\% against 0.216\%, 0.313\% and 0.192\%). We also tried an alternative, \revnew{\emph{Weighted NTP}}, which puts the concept loss's weight $\lambda$ on the observed word $w_0$ alone: compared with Conceptizer, it raises the observed word $w_0$'s probability further and reaches comparable downstream performance, but gives the contextual synonyms $w_1,\dots,w_m$ less probability and has a higher global perplexity.

On content-word positions, Randomized synonyms has the lowest perplexity in \rev{8 of 9} models, and Conceptizer in Falcon3 7B (Fig.~\ref{fig:ppl-matrix}). This lead comes from content words that split into several tokens. Randomized synonyms raises the probability of their first tokens that belong to many concepts (\emph{un-}, \emph{re-}) but gives the gain back on the rest of the word (seed 0; Appendix~\ref{app:controls}). Conceptizer comes at a modest cost in global perplexity, smaller than Randomized synonyms' in every model (for Llama 1B, 19.31 against 17.85 and 19.88; Fig.~\ref{fig:ppl-matrix}).

\subsection{Downstream Evaluations}
\label{sec:downstream}

\rev{To assess the impact of concept training on downstream performance, we use 3 task groups, each weighting its tasks equally} (splits and metrics in Appendix~\ref{app:configuration}). \textbf{Knowledge} is 4 science-question tasks, ARC-Challenge, ARC-Easy \citep{clark2018arc}, OpenBookQA \citep{mihaylov2018openbookqa} and SciQ \citep{welbl2017sciq}, scored by zero-shot answer likelihood \citep{gao2024harness}. \textbf{Sentiment} is 9 classification tasks of sentiment, emotion and toxicity from MTEB \citep{muennighoff2023mteb,enevoldsen2025mmteb}: IMDb \citep{maas2011imdb}, ToxicConversations \citep{jigsaw2019toxicity,borkan2019nuanced}, TweetSentimentExtraction \citep{maggie2020tweet}, Emotion \citep{saravia2018carer}, ToxicChat \citep{lin2023toxicchat}, PoemSentiment \citep{sheng2020poetry}, FrenkEn \citep{ljubesic2019frenk}, YelpReviewFull \citep{zhang2015charcnn} and AmazonPolarity \citep{mcauley2013hidden,zhang2015charcnn}. A logistic-regression classifier trained on 8 to 128 labeled examples per label measures how well final-layer hidden states separate the labels. \textbf{Commonsense} is HellaSwag \citep{zellers2019hellaswag}, COPA \citep{roemmele2011copa} and WinoGrande \citep{sakaguchi2020winogrande}, \rev{scored like knowledge}.

\paragraph{Results.}
\textbf{On \rev{knowledge tasks}, Conceptizer has the best score against the 3 baselines in 8 of 9 models and outperforms them in 26 of 27 comparisons} (9 models $\times$ 3 baselines; Fig.~\ref{fig:downstream}; Table~\ref{tab:seed-knowledge}). Its gains over \rev{NTP post-training}, 0.54 to 1.90 points, have paired training-seed intervals (pointwise 95\%, Student-$t$) above zero in all models but Llama 8B. Its gains over Randomized synonyms and the base model have intervals above zero in 7 models each. Per task, Conceptizer is best on OpenBookQA in all 9 models and on ARC-Challenge and ARC-Easy in 6 each (Table~\ref{tab:tasks-knowledge}).

\textbf{On \rev{sentiment tasks}, Conceptizer has the best score against the 3 baselines in 8 of 9 models and outperforms them in 26 of 27 comparisons} (Fig.~\ref{fig:downstream}; Table~\ref{tab:seed-sentiment}). Its gains over \rev{NTP post-training} (0.46 to 1.01 points) and Randomized synonyms (0.78 to 1.74 points) have paired intervals above zero in all 9 models. Its gains over the base model have intervals above zero in 6 models. Per task, Conceptizer is best on TweetSentimentExtraction in all 9 models, on the 7 other tasks in 5 to 7 models each, and on YelpReviewFull in none (Table~\ref{tab:tasks-sentiment}).

\textbf{On \rev{commonsense tasks}, the methods are comparable: Conceptizer, \rev{NTP post-training} and the base model each have the best score in 3 models} (Fig.~\ref{fig:downstream}; Table~\ref{tab:seed-commonsense}). Conceptizer is within 1.80 points of the best and above Randomized synonyms in every model. Against \rev{NTP post-training}, its paired intervals include zero \rev{except in} Llama 1B ($+1.78$ points, all from COPA's 100 \rev{items}).

\section{Conclusion}We introduced Conceptizer, a training framework that rewards meaning across the whole next-token distribution. Its fixed, model-agnostic concept vocabulary is built once from contextual synonyms and shared by every model through concept memberships, which standardizes the concept curation pipeline. Because earlier concept losses are special cases of its concept loss, Conceptizer also standardizes the concept training process. Against 3 baselines, across 3 model families and 3 scales, it gives the observed word and its contextual synonyms the most probability at held-out concept-target positions. It also leads on \rev{knowledge} and \rev{sentiment} tasks and remains comparable on \rev{commonsense}, at a marginal cost in global perplexity.

\section*{Limitations}The candidate generators and the judge are open models, not frontier models, and are more prone to mistakes. \rev{Biases of these models and of WordNet can also pass into the synonym pools and the concepts built from them.} First-token memberships can also link half-related words, such as \emph{care} and \emph{carelessness}. The judged-synonym probability is measured against synonyms kept by the same judge, so it is a reference point for how the training signal generalizes, not ground truth about meaning. By design, the concept target $\mathbf q_i$ can ask more of the closest concept $c^\star$ than \rev{the model concept distribution $\mathbf p_{\mathrm c,i}$} can give, although \rev{the concept loss $\mathcal L_{\mathrm{concept}}$ still asks to raise} \revnew{the logits of} the tokens whose concepts are under-served on average (Appendix~\ref{app:math}). How much of each word's share \rev{$\rho_j$} to send directly to the closest concept $c^\star$ ($\sigma_j$) deserves more careful experiments in future work. The hyperparameters follow \revnew{a sweep} (Appendix~\ref{app:sensitivity}) and the settings of previous work, and could be tuned more thoroughly. \rev{Concept-target positions} are only 0.15 to 0.24\% of training positions: about a third of positions are content words, only 3\% of those are collected for labeling (Llama), and a target needs its observed word $w_0$ and at least one synonym to be single tokens, which most collected occurrences are not (Appendix~\ref{app:data-eligibility}). Words that split into several tokens already have synonym pools and judge weights $r_j$, so expanding Conceptizer to multi-token words will unlock more coverage in future work without extra data curation. We study English text and models up to 8B parameters, which leaves larger models to future work.
\section*{Ethical Considerations}
This work post-trains released LMs (Llama, Qwen2.5 and Falcon3) on public web text and builds its concept vocabulary with other released models (Llama-3.3-70B-Instruct and Qwen2.5-72B-Instruct, with gpt-oss-120B as the judge); all of these models may carry biases from their training data, which the post-trained models and the synonym pools can inherit. AI assistants supported research planning, implementation, code review, literature retrieval, figure preparation and manuscript editing; the authors made the scientific decisions, verified the sources and are responsible for the paper.
\bibliography{references/references}
\clearpage\appendix
\makeatletter
\setlength{\@fptop}{0pt}\setlength{\@fpsep}{12pt plus 4pt}\setlength{\@fpbot}{0pt plus 1fil}
\setlength{\@dblfptop}{0pt}\setlength{\@dblfpsep}{12pt plus 4pt}\setlength{\@dblfpbot}{0pt plus 1fil}
\makeatother
\renewcommand{\topfraction}{0.92}\renewcommand{\dbltopfraction}{0.92}\renewcommand{\textfraction}{0.06}
\renewcommand{\floatpagefraction}{0.75}\renewcommand{\dblfloatpagefraction}{0.75}
\setcounter{topnumber}{3}\setcounter{dbltopnumber}{3}\setcounter{totalnumber}{4}
\section{How the Concept Loss Acts on Each Word}
\label{app:walkthrough}
Appendix~\ref{app:walkthrough} follows the concept loss $\mathcal L_{\mathrm{concept}}$ at the running example, after ``…practice the following sun'', for \rev{Llama-3.2-1B-Instruct before the first optimizer step, when the model is still the base model}. Table~\ref{tab:concept-loss-walkthrough} gives the quantities of Section~\ref{sec:method} for each kind of word. The pulls in Table~\ref{tab:concept-loss-walkthrough} are $\sum_c\omega_{vc}\,q_i(c)/p_{\mathrm c,i}(c)$, the sum of the token $v$'s uplift shares over all its concepts $\mathcal H(v)$\revnew{, with the membership weights $\omega_{vc}$, the concept target $\mathbf q_i$ and the model concept distribution $\mathbf p_{\mathrm c,i}$. The} pushes are $p_{\mathrm w,i}(v)\,(\text{pull}-1)$\revnew{, where $\mathbf p_{\mathrm w,i}$ is the next-token distribution. A push is the change the concept loss asks for in the logit $z_i(v)$, the negative of the gradient in Equation~\ref{eq:gradient-limit}. Table~\ref{tab:concept-loss-walkthrough} gives it before the objective $\mathcal L$ weights each concept target $\mathbf q_i$ by $\lambda/|\mathcal S|$, the concept-loss weight over the number of concept-target positions (Equation~\ref{eq:loss}).} \revnew{A concept's uplift is its target divided by its model probability, $q_i(c)/p_{\mathrm c,i}(c)$, and a token's logit $z_i(v)$ rises when its push is positive, that is, when the concept loss raises it, and falls when its push is negative.}

\begin{table}[!ht]\footnotesize\setlength{\tabcolsep}{2.5pt}
\noindent\begin{tabular*}{\linewidth}{@{\extracolsep{\fill}}lrrrrr@{}}
\multicolumn{6}{@{}l}{\textbf{(a) Member words of the closest concept $c^\star$, \#1450}}\\
\multicolumn{6}{@{}l}{$p_{\mathrm c,i}(c_{1450})=5.23\%$, $q_i(c_{1450})=68.7\%$: uplift 13.1}\\
\toprule
 & & & Uplift & & \\
Word & Concepts & $p_{\mathrm w,i}$ & share & Pull & Push\\
$v$ & $|\mathcal H(v)|$ & (\%) & ($\times$) & ($\times$) & (logit)\\
\midrule
\multicolumn{6}{@{}l}{\emph{Fewer than 14 concepts}}\\
safety & 9 & 32.1 & 1.46 & 1.58 & $+0.19$\\
security & 9 & 0.07 & 1.46 & 1.57 & $+0.0004$\\
fortress & 4 & $<$0.001 & 3.28 & 3.28 & $+2{\times}10^{-7}$\\
bunker & 9 & $<$0.001 & 1.46 & 1.48 & $+3{\times}10^{-8}$\\
\addlinespace[4pt]
\multicolumn{6}{@{}l}{\emph{14 or more concepts}}\\
protection & 22 & 32.1 & 0.60 & 1.06 & $+0.02$\\
shield & 53 & 0.014 & 0.25 & 0.30 & $-0.0001$\\
\bottomrule
\end{tabular*}

\bigskip
\noindent\begin{tabular*}{\linewidth}{@{\extracolsep{\fill}}llrrrr@{}}
\multicolumn{6}{@{}l}{\textbf{(b) Other concepts in the target}}\\
\toprule
 & & & & & Uplift\\
Concept & Word & $p_{\mathrm c,i}(c)$ & $q_i(c)$ & Uplift & share\\
$c$ & $v$ ($|\mathcal H(v)|$) & (\%) & (\%) & ($\times$) & ($\times$)\\
\midrule
\multicolumn{6}{@{}l}{\emph{Under-served}}\\
\#1379 & prevention (8) & 0.160 & 1.447 & 9.0 & 1.13\\
\#2859 & care (45) & 0.022 & \revnew{0.091} & 4.2 & 0.09\\
\addlinespace[4pt]
\multicolumn{6}{@{}l}{\emph{Over-served}}\\
\#5336 & protection (22) & \revnew{5.042} & \revnew{0.811} & 0.16 & 0.007\\
\bottomrule
\end{tabular*}

\bigskip
\noindent\begin{tabular*}{\linewidth}{@{\extracolsep{\fill}}lrrrrr@{}}
\multicolumn{6}{@{}l}{\textbf{(c) Words with no target}}\\
\toprule
 & & & Uplift & & \\
Word & Concepts & $p_{\mathrm w,i}$ & share & Pull & Push\\
$v$ & $|\mathcal H(v)|$ & (\%) & ($\times$) & ($\times$) & (logit)\\
\midrule
\multicolumn{6}{@{}l}{\emph{All concepts outside the target}}\\
protective & 7 & 1.8 & none & 0 & $-0.018$\\
\addlinespace[4pt]
\multicolumn{6}{@{}l}{\emph{One over-served concept in the target}}\\
habits & 11 & 2.1 & none & 0.04 & $-0.020$\\
\addlinespace[4pt]
\multicolumn{6}{@{}l}{\emph{In no concept}}\\
-related & \rev{1 ($\bot$)} & 6.3 & none & 0 & $-0.063$\\
\bottomrule
\end{tabular*}
\caption{\textbf{What the concept loss $\mathcal L_{\mathrm{concept}}$ asks of each word at the running example}, before the first optimizer step, where the model is the base model. $p_{\mathrm w,i}$ and $p_{\mathrm c,i}$: \rev{the next-token distribution and the model concept distribution} (Equation~\ref{eq:projection}), Fig.~\ref{fig:concept-train}'s $p$; $q_i$: the concept target (Equation~\ref{eq:target}), Fig.~\ref{fig:concept-train}'s $q$; Concepts: the word's number of concepts $|\mathcal H(v)|$ (in parentheses in panel b); Uplift share: the uplift of \#1450 (panels a and c) or of the row's concept (panel b), divided by the word's number of concepts $|\mathcal H(v)|$. Words with no target: neither the observed word $w_0$ nor one of its contextual synonyms $w_1,\dots,w_m$. Panel (a) splits at 14, the first whole number above the uplift 13.1. Pushes are on logits $z_i(v)$, not measured changes.}\label{tab:concept-loss-walkthrough}
\end{table}

\paragraph{What the cases show.}
\textbf{The whole concept is pulled, whether or not a member word fits the sentence.} \revnew{The logits of} \emph{fortress} (4 concepts) and \emph{bunker} (9), which do not fit \rev{the sentence}, rise with \revnew{those of} \emph{security} and the contextual synonyms $w_1,\dots,w_m$.

\textbf{\revnew{The logits of tokens} with fewer concepts than the uplift rise.} For a token of concept \#1450 with fewer than 13.1 concepts, \#1450 alone adds more than 1 to its pull, so \revnew{the logits of} all 23 such tokens rise. \revnew{The other 40 logits all fall except that of} \emph{protection}, a contextual synonym $w_1$ whose 21 other concepts also get part of the target $\mathbf q_i$ (at least \revnew{0.811}\% each, as \#5336 in panel~b) and lift its pull from 0.60 to 1.06.

\textbf{A contextual synonym $w_j$ is pulled through all its concepts $\mathcal H(w_j)$.} The 6 concepts only \emph{prevention} feeds each get 1.447\% of the target $\mathbf q_i$ but only \revnew{0.157 to 0.217}\% \rev{of the model concept distribution $\mathbf p_{\mathrm c,i}$}, so together they add 6.17 to its pull, almost 4 times what \#1450 adds.

\textbf{\revnew{The logit of a} synonym outside the closest concept $c^\star$ still rises.} \emph{Care} sends none of its 4.1\% \revnew{judge-weight} share \rev{$\rho_3$} to \#1450 and \revnew{0.091}\% to each of its 45 concepts (\#3086, shared with \emph{protection}, gets $\revnew{0.811+0.091=0.902}\%$). The base model's \revnew{next-token distribution $\mathbf p_{\mathrm w,i}$ puts} 0.97\% on \emph{care}\revnew{, which} gives each only about 0.022\% \revnew{of the model concept distribution $\mathbf p_{\mathrm c,i}$}: on \#2859 the target $\mathbf q_i$ asks \revnew{0.091}\% and gets 0.022\%, an uplift of 4.2, and \emph{care}'s pull is 3.39.

\textbf{Concepts the sentence rules out get nothing.} \revnew{The observed word $w_0$ assigns all of its share $\rho_0$ to the closest concept $c^\star$ ($\sigma_0=1$). Of \emph{safety}'s 8 other concepts (other than \#1450), the 5 that contain \emph{protection} get part of the target $\mathbf q_i$, but the 3 that contain no contextual synonym $w_j$ of this occurrence are outside it: \#6298 \{\emph{deliverance}, \emph{rescue}, \emph{salvation}, \dots\}, \#6387 \{\emph{lifelines}, \emph{ropes}, \emph{safetylines}, \dots\} and \#6389 \{\emph{aid}, \emph{safety-net}, \emph{support}, \dots\}.}

\textbf{\revnew{The logits of tokens} in no concept fall.} The pushes sum to zero (Appendix~\ref{app:math}), so every rise \revnew{in a logit} is paid for. Here the tokens in no concept, such as the word pieces \emph{-related} and \emph{-safe}, hold 20.9\% and \revnew{their logits} all fall\rev{, because} the concept target $\mathbf q_i$ gives the no-concept outcome $\bot$ nothing.

\textbf{\rev{First-token memberships add concepts.}} \emph{Care} joins 12 of its 45 concepts only through longer words that begin with its token, for example \emph{careening}, \emph{careworn}, \emph{carefree} and \emph{carelessness}.

\paragraph{The same example in the 3 families.}
Table~\ref{tab:three-family-example} repeats the running example in each family's own training file. \revnew{Its 4 words have the same numbers of concepts $|\mathcal H(v)|$ in all 3 tokenizers, except \emph{care}, which is not in \#1450. So the target $\mathbf q_i$ on \#1450 is the same in all 3, 68.7\%. Its \emph{bound}, the most probability the model concept distribution $\mathbf p_{\mathrm c,i}$ can put on \#1450 (Appendix~\ref{app:math}), is also 50\% in all 3. It is set by the tokens of \#1450 in just 2 concepts:} \emph{safeg} in Llama and Qwen2.5 (Appendix~\ref{app:data}), \emph{defences} and \emph{fortifications} in Falcon3. What differs is how each base model spreads its probability, which changes every uplift and pull. In each family, \revnew{the logit $z_i(v)$ of} every token of \#1450 with fewer concepts than \revnew{\#1450's} uplift rises\revnew{: 23 of 23 in Llama (uplift 13.1), 17 of 17 in Qwen2.5 (10.9) and 19 of 19 in Falcon3 (8.5).}

\revnew{\emph{Care} has 45 concepts in Llama and Qwen2.5 but 44 in Falcon3, through first-token memberships.} Falcon3 keeps \emph{carefree} and \emph{carelessness} as single tokens, so \emph{care} loses the 3 concepts it reaches only through them \revnew{(\#839 \{\emph{carelessness}, \emph{error}, \emph{mistake}, \dots\}, \#7049 \{\emph{carefree}, \emph{casual}, \emph{offhand}, \dots\} and \#7095 \{\emph{calm}, \emph{carefree}, \emph{relaxed}, \dots\})}. Falcon3 also splits \emph{caretaking} into \emph{care} and \emph{taking}, not \emph{caret} and \emph{aking} as in Llama and Qwen2.5, which gives \emph{care} 2 other concepts \revnew{(\#6682 \{\emph{caretaking}, \emph{nursing}, \emph{nurturing}, \dots\} and \#7111 \{\emph{caretaking}, \emph{maintenance}, \emph{repair}, \dots\})}.

\begin{table}[!ht]\footnotesize\setlength{\tabcolsep}{3pt}
\noindent\begin{tabular*}{\linewidth}{@{\extracolsep{\fill}}lrrr@{}}
\toprule
 & Llama & Qwen2.5 & Falcon3\\
\midrule
\multicolumn{4}{@{}l}{\emph{Tokenizer}}\\
Concepts per member token & 11.9 & 11.9 & 8.3\\
Tokens in \#1450 & 63 & 63 & 67\\
Bound on \#1450 (\%) & 50 & 50 & 50\\
\addlinespace[4pt]
\multicolumn{4}{@{}l}{\emph{Number of concepts $|\mathcal H(v)|$}}\\
safety & 9 & 9 & 9\\
protection & 22 & 22 & 22\\
prevention & 8 & 8 & 8\\
care & 45 & 45 & 44\\
\addlinespace[4pt]
\multicolumn{4}{@{}l}{\emph{Base model: $p_{\mathrm w,i}$ (\%)}}\\
safety & 32.1 & 49.9 & 65.5\\
protection & 32.1 & 16.2 & 12.9\\
prevention & 1.24 & 0.04 & 1.54\\
care & 0.97 & 0.34 & 0.29\\
$p_{\mathrm c,i}(c_{1450})$ (\%) & 5.23 & 6.29 & 8.12\\
Target $q_i(c_{1450})$ (\%) & 68.7 & 68.7 & 68.7\\
Uplift on \#1450 ($\times$) & 13.1 & 10.9 & 8.5\\
\addlinespace[4pt]
\multicolumn{4}{@{}l}{\emph{Base model: pull ($\times$)}}\\
safety & 1.58 & 1.31 & 1.02\\
protection & 1.06 & 1.33 & 1.42\\
prevention & 7.87 & 166 & 5.24\\
care & 3.39 & 10.4 & 11.3\\
\addlinespace[4pt]
\multicolumn{4}{@{}l}{\emph{After post-training: the 4 words' total $p_{\mathrm w,i}$ (\%)}}\\
\rev{NTP post-training} & 75.8 & 67.4 & 76.6\\
Conceptizer & 71.4 & 75.7 & 47.4\\
\bottomrule
\end{tabular*}
\caption{\textbf{\revnew{The running example keeps its target and its bound on \#1450 in all 3 families, while each base model changes the uplifts and pulls.}} The same training position with each family's different concept memberships $\mathcal H(v)$ (from its tokenizer) and different base model (Llama-3.2-1B-Instruct, Qwen2.5-1.5B-Instruct and Falcon3-1B-Instruct). Concepts per member token: the average number of concepts a token belongs to, $|\mathcal H(v)|$, over the tokens that belong to at least one concept. \rev{Bound on \#1450: the most probability the model concept distribution $\mathbf p_{\mathrm c,i}$ can put on \#1450 (Equation~\ref{eq:closest-bound}).} Symbols as in Table~\ref{tab:concept-loss-walkthrough}. The last 2 rows give the 4 words' combined probability under the seed-0 adapters at this training position, not held-out evidence.}\label{tab:three-family-example}
\end{table}

\section{Derivations for the Concept Loss}
\label{app:math}
\revnew{Appendix~\ref{app:math} derives the gradient of the concept loss through the concept memberships, the target on the closest concept, the most probability the model concept distribution can put on a concept, and earlier losses as special cases.} The implemented concept loss at position $i$ is $\ell_i=-\sum_{c\in\mathcal C^+}q_i(c)\log\big(p_{\mathrm c,i}(c)+\varepsilon\big)$ with $\varepsilon=10^{-12}$, which keeps the logarithm finite when a concept gets no probability. \rev{With $\varepsilon=0$, the mean of $\ell_i$ over the concept-target positions $\mathcal S$ is the concept loss $\mathcal L_{\mathrm{concept}}$ (Equation~\ref{eq:concept-loss}).} \rev{In an optimizer step with no concept-target positions, the concept loss is zero.}

\paragraph{Gradient through concept memberships.}
For output tokens $u,v\in\mathcal V$, the softmax gives $\partial p_{\mathrm w,i}(u)/\partial z_i(v)=p_{\mathrm w,i}(u)\big(\mathbf 1[u=v]-p_{\mathrm w,i}(v)\big)$, where $\mathbf 1[u=v]$ is 1 if $u=v$ and 0 otherwise, so by Equation~\ref{eq:projection}
\begin{equation}
 \frac{\partial p_{\mathrm c,i}(c)}{\partial z_i(v)}
 =p_{\mathrm w,i}(v)\big(\omega_{vc}-p_{\mathrm c,i}(c)\big).
 \label{eq:projection-gradient}
\end{equation}
Raising $z_i(v)$ thus raises $p_{\mathrm c,i}(c)$ only for the concepts whose membership weight $\omega_{vc}$ exceeds their current probability $p_{\mathrm c,i}(c)$. The exact derivative of the implemented loss $\ell_i$ is
\begin{equation}
 \frac{\partial\ell_i}{\partial z_i(v)}
 =p_{\mathrm w,i}(v)\sum_{c\in\mathcal C^+}q_i(c)\,
 \frac{p_{\mathrm c,i}(c)-\omega_{vc}}{p_{\mathrm c,i}(c)+\varepsilon}.
 \label{eq:gradient-exact}
\end{equation}
When $\varepsilon\to0$ and $p_{\mathrm c,i}(c)>0$ wherever $q_i(c)>0$, $\sum_c q_i(c)=1$ turns Equation~\ref{eq:gradient-exact} into
\begin{equation}
 p_{\mathrm w,i}(v)\Big[1-\sum_{c:\,q_i(c)>0}\frac{\omega_{vc}\,q_i(c)}{p_{\mathrm c,i}(c)}\Big].
 \label{eq:gradient-limit}
\end{equation}
The sum is the pull of $v$ (Section~\ref{sec:method}), the average uplift of $v$'s concepts $\mathcal H(v)$, since $v$'s probability is split evenly over them, and the push $p_{\mathrm w,i}(v)\,(\text{pull}-1)$ is minus this derivative. \rev{The loss $\ell_i$ therefore asks for $v$'s logit $z_i(v)$ to rise} exactly when its concepts are under-served on average. Written out, the push is \revnew{$v$'s \emph{routed target}, }the part of the target $\mathbf q_i$ routed to $v$\revnew{,} minus its probability $p_{\mathrm w,i}(v)$,
\begin{equation}
\begin{split}
 &p_{\mathrm w,i}(v)\,(\text{pull}-1)\\
 &\quad=\sum_{c:\,q_i(c)>0}q_i(c)\,\frac{\omega_{vc}\,p_{\mathrm w,i}(v)}{p_{\mathrm c,i}(c)}-p_{\mathrm w,i}(v),
\end{split}
 \label{eq:push-routed}
\end{equation}
where each concept's target $q_i(c)$ is split among its tokens in proportion to their current probability within the concept, $\omega_{vc}\,p_{\mathrm w,i}(v)/p_{\mathrm c,i}(c)$. \revnew{At the running example, \emph{safety} supplies $32.1\%/9=3.57\%$ of $p_{\mathrm c,i}(c_{1450})=5.23\%$, that is 68\%, so it is routed 68\% of the target $q_i(c_{1450})=68.7\%$, or 46.8\%, and 50.7\% in all once its 5 other concepts in the target are counted. Against its 32.1\%, that is the push of $0.507-0.321=+0.19$.} At a given step the concept loss $\ell_i$ therefore \rev{asks the logits $\mathbf z_i$ for the same change as} a token-level cross-entropy toward this routed target would, but the routed target changes with the model. Summing Equation~\ref{eq:push-routed} over $\mathcal V$, $\sum_v\omega_{vc}\,p_{\mathrm w,i}(v)=p_{\mathrm c,i}(c)$ leaves $\sum_c q_i(c)-1=0$: the pushes sum to zero, so every rise \revnew{in a logit} is paid for by falls elsewhere. With $\varepsilon>0$, the 1 in Equation~\ref{eq:gradient-limit} becomes $\sum_c q_i(c)\,p_{\mathrm c,i}(c)/(p_{\mathrm c,i}(c)+\varepsilon)$, slightly below 1, and each uplift becomes $q_i(c)/(p_{\mathrm c,i}(c)+\varepsilon)$.

\paragraph{The target on the closest concept.}
By Equation~\ref{eq:target}, the target's mass on the closest concept $c^\star$ is
\begin{equation}
 q_i(c^\star)=\sum_{j=0}^{m}\rho_j\,\kappa_j,\qquad
 \kappa_j=\sigma_j+(1-\sigma_j)\,\omega_{w_jc^\star},
 \label{eq:target-closest}
\end{equation}
where $\kappa_j$, the ratio of Fig.~\ref{fig:concept-train}, is the fraction of word $w_j$'s judge-weight share $\rho_j$ that reaches $c^\star$\rev{, directly ($\sigma_j$) or through the concept memberships of $w_j$'s token ($(1-\sigma_j)\,\omega_{w_jc^\star}$)}: 1 for the observed word $w_0$ ($\sigma_0=1$), $\tfrac12+\tfrac12/|\mathcal H(w_j)|$ for a synonym whose token \rev{belongs to} $c^\star$ ($\sigma_j=\tfrac12$; $|\mathcal H(w_j)|$ is the number of concepts of $w_j$'s token), and 0 for any other synonym ($\sigma_j=0$ and $\omega_{w_jc^\star}=0$).

\paragraph{\rev{The most probability the model concept distribution can put on a concept.}}
By Equation~\ref{eq:projection}, $\mathbf p_{\mathrm c,i}=\sum_v p_{\mathrm w,i}(v)\,\boldsymbol\omega_v$ is a probability-weighted average of the tokens' \rev{membership weights} $\boldsymbol\omega_v=(\omega_{vc})_{c\in\mathcal C^+}$, so \rev{$\mathbf p_{\mathrm c,i}$ never gives} a concept more than its largest membership weight, $p_{\mathrm c,i}(c)\le\max_v\omega_{vc}$, with equality only when all probability is on such \rev{tokens}. For the closest concept $c^\star$,
\begin{equation}
 p_{\mathrm c,i}(c^\star)\le\frac{1}{\min_{v:\,c^\star\in\mathcal H(v)}|\mathcal H(v)|},
 \label{eq:closest-bound}
\end{equation}
the inverse of the fewest concepts that any token of $c^\star$ belongs to, below 1 whenever every token of $c^\star$ also belongs to another concept. At the running example, every token of \#1450 belongs to at least 2 concepts, so \rev{the model concept distribution $\mathbf p_{\mathrm c,i}$ never gives} \#1450 more than 50\%, against the 68.7\% the target $\mathbf q_i$ asks (Table~\ref{tab:three-family-example}).\revnew{\footnote{The bound is on the concept, not on a token: \emph{safety}'s routed 50.7\% above is a target for its token, which gives \#1450 only 1/9 of its probability, so even all of the probability on \emph{safety} would give \#1450 just 11.1\%. Only the tokens of \#1450 in just 2 concepts, such as \emph{safeg} for Llama, can bring it to 50\% (Appendix~\ref{app:walkthrough}).}} \rev{With $\varepsilon=0$, $\ell_i$ is the target's entropy plus $D_{\mathrm{KL}}(\mathbf q_i\Vert\mathbf p_{\mathrm c,i})$, so the loss reaches its minimum, the target's entropy, only at $\mathbf p_{\mathrm c,i}=\mathbf q_i$.} A target that asks its closest concept $c^\star$ for more than the bound therefore keeps the divergence above zero for every model, and training \rev{can move toward the target but never reach it}. The bound concerns one coordinate, so meeting it does not make a target attainable. Many training targets ask their closest concept $c^\star$ for more than this bound: 57.9\% of Llama's, 58.2\% of Qwen2.5's and 36.4\% of Falcon3's.

\paragraph{Earlier losses as special cases.}
With $\varepsilon=0$, $\ell_i=-\sum_{c}q_i(c)\log p_{\mathrm c,i}(c)$ rests on 3 choices: the \rev{concept vocabulary} $\mathcal C$, each token's concepts $\mathcal H(v)$, which fix the membership weights $\omega_{vc}$, and the target $\mathbf q_i$. Let $T_i\subseteq\mathcal V$ be the tokens an earlier method rewards at position $i$. Three settings of these choices recover known losses.
\textbf{Next-token prediction.} Give every token a concept of its own, $\mathcal H(v)=\{v\}$, so $\omega_{vc}=\mathbf 1[c=v]$ and $p_{\mathrm c,i}(v)=p_{\mathrm w,i}(v)$, and put the whole target on the observed token $g_i$, $\mathbf q_i=\mathbf e_{g_i}$:
\begin{equation}
 \ell_i=-\log p_{\mathrm w,i}(g_i).
 \label{eq:special-ntp}
\end{equation}
\textbf{Next-concept prediction} \citep{iyer2026beyond}. Keep these one-token concepts and spread the target evenly over $T_i$, $\mathbf q_i=|T_i|^{-1}\sum_{t\in T_i}\mathbf e_t$:
\begin{equation}
 \ell_i=-\frac{1}{|T_i|}\sum_{t\in T_i}\log p_{\mathrm w,i}(t),
 \label{eq:special-iyer}
\end{equation}
the negative of their next-concept-prediction objective (their Section~2.2.2), which rewards each token of $T_i$ on its own. The match is exact when every completion is a single token: their loss also scores multi-token completions over their whole token span (their Appendix~D), which a loss at one position cannot express. Their data augmentation, one NTP example per token of $T_i$, averages to the same loss.
\textbf{Pooled synonyms} \citep{zhang2026concepts}. Let $T_i$ hold the observed word $w_0$ and its contextual synonyms $w_1,\dots,w_m$, give them one concept $c_T$ of their own, $\mathcal H(t)=\{c_T\}$ for every $t\in T_i$, so $\omega_{tc_T}=1$ and $p_{\mathrm c,i}(c_T)=\sum_{t\in T_i}p_{\mathrm w,i}(t)$, and put the whole target on it, $\mathbf q_i=\mathbf e_{c_T}$:
\begin{equation}
 \ell_i=-\log\sum_{t\in T_i}p_{\mathrm w,i}(t),
 \label{eq:special-zhang}
\end{equation}
their concept loss (their Section~3.3), which rewards the pooled probability.
The identities concern the \rev{concept loss} alone: \citeauthor{zhang2026concepts} interpolate it with NTP at content words and switch it off once $T_i$ holds 60\% of the probability, while Conceptizer adds it to NTP and retention, $\mathcal L_{\mathrm{NTP}}$ and $\mathcal L_{\mathrm{retention}}$ (Equation~\ref{eq:loss}).

\section{Data \texorpdfstring{\rev{Preparation}}{Preparation} Details}
\label{app:data}
\revnew{Appendix~\ref{app:data}} gives the exact rules of Section~\ref{sec:data} in the order the data pass through them. Occurrences of content words are collected, and their synonym candidates are proposed and judged. A sample of the judged occurrences builds the concept vocabulary, which concept memberships link to each tokenizer. Each occurrence \revnew{that passes 6 rules} then gets a concept target inside the training and held-out files.

\paragraph{Collection.}
Occurrences come from English-source FineWeb-Edu documents (release v1.4.0, its sample-100BT subset) with a language score of at least 0.90 (FineWeb-Edu's fastText confidence that a document is English) and at most 1,500,000 characters. Heuristic math and code filters keep the judged sentences to ordinary prose. They apply to each document's first 2,000 characters and to each sentence, and exclude text shorter than 20 characters, with more than 25\% non-alphabetic, non-space characters or more than 15\% digits, or matching code or LaTeX patterns. The filters act on collection only, so training windows, taken from the full documents, can still contain math or code. spaCy 3.8.14 with \texttt{en\_core\_web\_sm} 3.8.0 tags the content words: alphabetic non-stopword nouns, verbs and adjectives, which leaves out proper nouns. A word is its lowercased spaCy lemma, so \emph{Edible} and \emph{edible} are one word.

\paragraph{Candidates.}
The 3 sources of Section~\ref{sec:data} are set \revnew{to cast a wide net}, since the judge removes what does not fit. Llama-3.3-70B-Instruct reads the sentence with the first occurrence of the word's spelling marked. Qwen2.5-72B-Instruct sees no sentence, so it responds once per word, and the response is reused for all of the word's occurrences (prompts in Appendix~\ref{app:prompts}). Both draw 5 samples at temperature 1.0 and seed 0, with at most 128 and 192 tokens per response, respectively. \revnew{Additionally,} WordNet \citep{miller1995wordnet}, read through NLTK \citep{bird2009nltk}, adds the \revnew{lemma names (synonyms) of all the word's senses} under its POS, without multiword phrases. \revnew{It has such synonyms for 20.7\% of the collected words, which hold 31.4\% of the occurrences.} The lowercased union of the 3 sources removes duplicates and records each candidate's sources. Falcon3's family proposed no candidates, yet in all 3 Falcon3 models Conceptizer has the lowest concept-target perplexity and the best knowledge score against the 3 baselines, and the best sentiment score in 2 (Fig.~\ref{fig:ppl-matrix}; Tables~\ref{tab:seed-knowledge} and~\ref{tab:seed-sentiment}).

\paragraph{The judge.}
The LLM judge, gpt-oss-120B, narrows the candidates to the contextual synonyms. It reads the same marked sentence with the word, its POS and the candidates in alphabetical order (prompt in Appendix~\ref{app:prompts}), at temperature 0.7 with high reasoning effort\revnew{, 3 times per occurrence. An occurrence is removed when at least 2 of its 3 answers mark the word as a name or code.} Table~\ref{tab:safety_judgments} lists the judge's answers for the 25 occurrences of \emph{safety}, with the judge weights of Equation~\ref{eq:judge-weight}.

\begin{table*}[tp]\centering\footnotesize\setlength{\tabcolsep}{3.8pt}\renewcommand{\arraystretch}{1.08}
\begin{tabular}{@{}rlrcccc@{}}
\toprule
 & Sentence (excerpt) & Candidates & \emph{protection} & \emph{prevention} & \emph{care} & \emph{security} \\
\midrule
1 & …child \emph{safety} seats and bicycle helmets & 39 & 3/3 & not proposed & 0/3 & 3/3 \\
2 & …follow certain \emph{safety} precautions & 44 & 3/3 & 0/3 & 0/3 & 3/3 \\
3 & …highway \emph{safety} was a top concern & 39 & 3/3 & not proposed & 0/3 & 3/3 \\
4 & …concerns about pilot and bird \emph{safety} & 39 & 3/3 & not proposed & 0/3 & 3/3 \\
5 & For the \emph{safety} of the birds, we cannot… & 41 & 3/3 & not proposed & 1/3 & 3/3 \\
6 & …related to \emph{safety} and herbal medicines & 41 & 3/3 & not proposed & 0/3 & 3/3 \\
7 & …\emph{safety} assessment of herbal medicines & 43 & 2/3 & not proposed & 0/3 & 2/3 \\
8 & …for \emph{safety} assessment apply to herbal… & 41 & 2/3 & not proposed & 0/3 & 2/3 \\
9 & …\emph{safety} assessment exist in their country & 41 & 3/3 & not proposed & 0/3 & 3/3 \\
10 & Regulatory requirements for \emph{safety}… & 41 & 1/3 & not proposed & 0/3 & 1/3 \\
11 & …requirements for \emph{safety} assessment of… & 41 & 1/3 & not proposed & 0/3 & 1/3 \\
12 & …requirements for \emph{safety} assessment. & 41 & 1/3 & not proposed & 0/3 & 0/3 \\
13 & …for \emph{safety} assessment detailed above & 41 & 2/3 & not proposed & 0/3 & 2/3 \\
14 & …mechanism for \emph{safety} requirements & 40 & 3/3 & not proposed & 0/3 & 3/3 \\
15 & …follow good summer \emph{safety} habits… & 43 & 3/3 & 2/3 & 0/3 & 3/3 \\
16 & …but \emph{safety} comes first when enjoying… & 41 & 3/3 & not proposed & 0/3 & 3/3 \\
17 & Keep the following \emph{safety} precautions… & 47 & 3/3 & 0/3 & 0/3 & 3/3 \\
\textbf{18} & \textbf{…practice the following sun \emph{safety} tips…} & \textbf{43} & \textbf{3/3} & \textbf{2/3} & \textbf{1/3} & \textbf{0/3} \\
19 & …Practice good summer \emph{safety} & 43 & 3/3 & 1/3 & 0/3 & 2/3 \\
20 & Talk to your kids about school \emph{safety} & 39 & 3/3 & not proposed & 0/3 & 3/3 \\
21 & Contemporary \emph{safety} principles of… & 40 & 3/3 & 1/3 (added) & 0/3 & 3/3 \\
22 & …multiple layers of \emph{safety} systems… & 40 & 3/3 & not proposed & 0/3 & 3/3 \\
23 & …capabilities and \emph{safety} profiles than… & 39 & 2/3 & not proposed & 0/3 & 0/3 \\
24 & …to meltdown and operating \emph{safety} & 39 & 3/3 & not proposed & 0/3 & 2/3 \\
25 & PIUS (process inherent ultimate \emph{safety}) & 39 & 3/3 & not proposed & 0/3 & 3/3 \\
\midrule
 & Kept in / proposed in, $k_j/n_j$ &  & 25/25 & 4/6 & 2/25 & 22/25 \\
 & Judge weight $r_j=(k_j+1)/(n_j+2)$ &  & 26/27 $=$ 0.963 & 5/8 $=$ 0.625 & 3/27 $=$ 0.111 & 23/27 $=$ 0.852 \\
\bottomrule\end{tabular}
\caption{\textbf{The judge's answers behind the running example.} The 25 collected occurrences of \emph{safety} and the number of candidates the 3 sources proposed. Each cell counts how many of the judge's 3 answers kept the word. Not proposed marks a word that was not a candidate, and (added) the one occurrence where the judge added \emph{prevention} itself. Bold: the running example of Figs.~\ref{fig:concept-train} and~\ref{fig:word-example}. Rows 11 and 12, and 21 and 22, are 2 occurrences of \emph{safety} in one sentence each, judged separately.}\label{tab:safety_judgments}
\end{table*}

\paragraph{Concept-construction sample.}
The concept vocabulary is built from a fixed sample of judged occurrences, the \revnew{\emph{concept-construction sample}}, chosen to hold words with several senses. It takes words with at least 15 occurrences in which the judge kept a candidate, and whose synonym pools form at least 2 groups when pools with a Jaccard similarity of at least 0.25 are linked. \revnew{Words with the most such occurrences enter first, with ties broken by file order, until the sample holds 150,000 occurrences; all 6,000 that enter have the maximum of 25.} \revnew{Their sentences are tokenized and truncated at 256 tokens; the 81 occurrences in which no token of the sentence covers the word are dropped, and the other 149,919, \revnew{from 17,326 documents,} form the concept-construction sample.} Two alignment checks then leave out 2,262 occurrences whose tokens do not spell the word and 2,423 that open their sentence. \revnew{Finally, occurrences in 60\% of the 17,326 documents build the concepts, and the other 40\% are held out from construction.} A word with fewer than 5 remaining occurrences in the 60\% builds no concept. This leaves the 88,367 occurrences of 5,939 words of Section~\ref{sec:data}.

\paragraph{Senses and concepts.}
\label{app:data-concepts}
Section~\ref{sec:data} builds the concepts in 2 steps. First, within a word, average linkage groups synonym pools into senses by their Dice similarity, comparing kept words after folding inflections (WordNet \texttt{morphy} under the word's POS, then the other POS, then suffix rules for \texttt{-ies}, \texttt{-ing}, \texttt{-ed} and plurals), so \emph{rising} and \emph{grown} compare as \emph{rise} and \emph{grow}. A sense's final pool keeps the surface forms: its occurrences' kept words and the word as written in each occurrence.

Second, across words, 2 senses form a \revnew{\emph{candidate pair}} when their final pools share at least 2 surface forms and reach the Jaccard threshold. Pairs are taken in descending similarity, and complete linkage merges 2 groups only when every pair between them qualifies, which keeps chains of partial overlaps from joining unrelated words. Senses of the same word stay in different concepts, since they are different meanings of that word. Only words in at most 200 final pools are used to look up candidate pairs, which keeps the search for pairs tractable, though every word counts in the similarity. Word lookup applies Unicode normalization, folds dash variants and removes case duplicates within a concept.

Concepts are numbered \#0 to \#9060 by first appearance in the concept-construction sample, and they have 2 to 169 words (median 20). Concept \#1450 merges a sense of \emph{bulwark} (11 occurrences, 44 words) with one of \emph{defence} (14 occurrences, 53 words), whose final pools share 25 of their 72 words, among them \emph{protection} and \emph{shield}. It also contains \emph{safety}, kept for \emph{defence} in 3 occurrences, and loose member words such as \emph{moat}, kept in 1. Appendix~\ref{app:sensitivity} varies the Dice and Jaccard thresholds and the linkage.

\paragraph{Concept memberships.}
Each tokenizer reaches the concepts through concept memberships (Section~\ref{sec:data}). A word's token is the first token of the word written with a leading space, as it appears after another word \revnew{(for Llama, \emph{safety} is one token, \texttt{\textvisiblespace safety}, but 2, \emph{s} and \emph{afety}, without the space)}, and the single-token checks use the same form. The 49,387 distinct member words \revnew{in the concept vocabulary} give 17,938 \revnew{\emph{member tokens}} for Llama, 17,870 for Qwen2.5 and 27,430 for Falcon3, because many words share a first token: \emph{care}'s token begins 8 member words (\emph{care}, \emph{careen}, \emph{careened}, \emph{careening}, \emph{careens}, \emph{carefree}, \emph{carelessness} and \emph{careworn}). All 9,061 concepts have a member token in each family, and a single-token member word in 8,926 concepts for Llama, 8,926 for Qwen2.5 and 9,029 for Falcon3.

\paragraph{Closest concept $c^\star$.}
\leavevmode\revnew{Each concept target \revnew{$\mathbf q_i$} sends the observed word \revnew{$w_0$}'s whole share \revnew{$\rho_0$} to its closest concept $c^\star$ (Section~\ref{sec:data}).} Among the concepts that list the observed word $w_0$, $c^\star$ is the one containing the most contextual synonyms $w_1,\dots,w_m$. \revnew{If several concepts tie for the most, the lowest-numbered one becomes $c^\star$. Ties occur in 33\% of Llama's and Qwen2.5's training targets and 27\% of Falcon3's: at the running example, \#1450 and \#2628 each contain 2 synonyms, and the lower-numbered \#1450 becomes $c^\star$.} The same occurrence gets the same closest concept $c^\star$ from Llama's and Qwen2.5's tokenizers in 99.98\% of the training targets they share and from all 3 tokenizers in \revnew{90.06}\%. Falcon3 keeps more synonyms as single tokens, which can move $c^\star$: \emph{citizens} takes \#1571 for Llama and Qwen2.5 but \#2203 for Falcon3, which also keeps \emph{citizenry} as one token. \revnew{The next paragraph covers 2 special cases in which counting contextual synonyms cannot choose $c^\star$.}

\paragraph{\rev{Token owners.}}
\rev{In 2 special cases, counting contextual synonyms $w_1,\dots,w_m$ cannot choose the closest concept $c^\star$. In the \revnew{\emph{first special case}}, $w_0$ is listed in several concepts, but none of them contains a contextual synonym (7\% of Llama's and Qwen2.5's training targets, 6\% of Falcon3's). In the \revnew{\emph{second special case}}, \revnew{$w_0$ is not listed in any concept}, and its token belongs to concepts only as the first token of longer member words (0.3\%, 0.3\% and 0.2\%). For example, no concept lists \emph{uninstall}, but \#6852 lists \emph{uninstalling}, which begins with the token of \emph{uninstall} in all 3 tokenizers.}

\rev{For the 2 special cases, each member token $v$, the first token of at least one member word, has an \emph{owner}: one of its concept memberships $\mathcal H(v)$, designated in advance. In the first special case, the owner of $w_0$'s token becomes the closest concept $c^\star$ if that owner lists $w_0$. Otherwise, the lowest-numbered concept that lists $w_0$ becomes $c^\star$. In the second special case, the owner of $w_0$'s token always becomes $c^\star$\revnew{, and rule 6 of ``Which occurrences get a target'' decides whether the target is kept}. For example, the token of \emph{uninstall} belongs only to \#6852, through \emph{uninstalling}, so \#6852 owns it and becomes $c^\star$ for every occurrence of \emph{uninstall}.}

\rev{Owners are set by 2 rules\revnew{, applied in order}. \revnew{The first rule gives each token to the lowest-numbered of its concept memberships $\mathcal H(v)$. However, alone, it would leave a concept owning no token whenever all of its member tokens also belong to lower-numbered concepts, and the second special case could never choose that concept. For example, the first rule gives the tokens of \emph{ages} and \emph{years} to \#265 \{\emph{aeon}, \emph{age}, \emph{ages}, \emph{centuries}, \dots, \emph{years}\}, leaving \#267 \{\emph{ages}, \emph{years}\} owning none.}}

\rev{\revnew{Therefore, the second rule lets each concept left owning no token take one of its member tokens from that token's owner under the first rule, as long as that owner keeps at least one other token. In the example, \#267 takes the token of \emph{years} from \#265. In the end, \#267 owns 1 token, and \#265 owns 10 for Llama. The first rule gave 21 tokens to \#265, and the second rule then moved 11 of them to 11 different concepts such as \#267, one each.}}

\rev{The second rule changes the owner of 25.9\% of Llama's member tokens, 26.1\% of Qwen2.5's and 12.7\% of Falcon3's. \revnew{After both rules, every concept is the owner of at least one token except 126 of the 9,061 (1.4\%) for Llama, 127 (1.4\%) for Qwen2.5 and 38 (0.4\%) for Falcon3, because each of their member tokens is its owner's only token.} \#3248 \{\emph{gullet}, \emph{larynx}, \emph{pharynx}, \emph{throat}\} is one: each of its 4 member tokens is the only token of a different owner, such as \#433 \{\emph{airway}, \emph{throat}, \emph{trachea}, \emph{windpipe}, \dots\}, which owns only the token of \emph{throat}. Owners matter only in the 2 special cases: every concept, owned or not, gets probability through all of its member tokens, and a concept that owns no token can still become $c^\star$ by counting contextual synonyms.}

\paragraph{Which occurrences get a target.}
\label{app:data-eligibility}
The concept loss reads one next-token distribution, so a concept target needs an observed word and synonyms that are single tokens, and a closest concept that contains the token that occurs. A collected occurrence therefore gets a concept target only if it passes 6 rules. \revnew{The first 3 check the occurrence, and the last 3 build and check the target.}

\revnew{\textbf{(1) Single tokens, not a name or code.}} The observed word $w_0$ is a single whole-word token, at most 1 of the judge's 3 answers marks it as a name or code, and at least one kept synonym is a single token.

\revnew{\textbf{(2) Concept memberships.}} The observed token $w_0$ belongs to a concept with fewer than 200 member words, and a single-token synonym belongs to a concept.

\revnew{\textbf{(3) Whole words, no negating prefix.}} The observed word $w_0$ starts and ends at word boundaries in the document, so its token is the whole word as written, and does not follow a negating prefix joined by a hyphen or dash, which would invert its meaning (\emph{non-judgmental} never gives \emph{judgmental} a target).

\revnew{\textbf{(4) At most 10 synonyms.}} Multi-token synonyms are dropped, and the rest are ranked by how many answers kept them and limited to 10 before judge weights $r_j$ are attached.

\revnew{\textbf{(5) The observed token in $c^\star$.}} The exact observed token $w_0$ at that position belongs to the closest concept $c^\star$, so the target's largest share lands on a concept that the observed token gives probability to.

\revnew{\textbf{(6) Lexical support.} In the second special case of the token owners, $w_0$ is listed in no concept, and the owner of its token becomes $c^\star$ only because that token is the first token of a member word. If no synonym is listed in $c^\star$ either, the target needs \emph{lexical support}: a synonym with the same WordNet lemma as a member word of $c^\star$, neither word having any other lemma, and no lemma in common with $w_0$. For example, the token of baking \emph{flour} is the first token of \emph{flourishes}, so in Llama and Qwen2.5 it is owned by an outbreak concept (\#6109 \revnew{\{}\emph{bloom}, \emph{burst}, \emph{flourishes}, \dots\revnew{\}}). None of \emph{flour}'s synonyms (\emph{cover}, \emph{dust}, \emph{powder}, \emph{coat}, \emph{sprinkle}) has the lemma of a member of \#6109, so the target is removed. This rule removes about 9 in 10 such occurrences (Table~\ref{tab:eligibility_funnel}), but it does not certify the sense of those it keeps.}

A training target ends with 1 to 10 contextual synonyms $w_1,\dots,w_m$, a median of 5 for Llama and Qwen2.5 and 6 for Falcon3: 1 synonym in 10.4\% of Llama's targets, 10.5\% of Qwen2.5's and 7.3\% of Falcon3's, and the cap of 10 in 15.0\% of Llama's, 15.0\% of Qwen2.5's and 23.6\% of Falcon3's. A synonym whose token belongs to no concept stays in the target $\mathbf q_i$, and its whole share $\rho_j$ goes to the no-concept outcome $\bot$. Table~\ref{tab:eligibility_funnel} counts each rule's removals.
\begin{table*}[tp]\centering\footnotesize\setlength{\tabcolsep}{4.0pt}\renewcommand{\arraystretch}{1.08}
\begin{tabular}{@{}lrrr@{}}
\toprule
 & Llama & Qwen2.5 & Falcon3 \\
\midrule
Judged occurrences, all documents & 2,327,384 & 2,327,384 & 2,327,384 \\
\quad outside the training documents' windows & 236,040 & 236,001 & 236,106 \\
Located in training documents' windows & 2,091,344 & 2,091,383 & 2,091,278 \\
\quad in discarded short document ends & 2,759 & 2,735 & 2,802 \\
In kept training windows & 2,088,585 & 2,088,648 & 2,088,476 \\
\addlinespace[2pt]
\multicolumn{4}{@{}l}{\emph{Removed, in the order the checks apply}} \\
\quad Observed word $w_0$ splits into several tokens & 1,598,264 & 1,599,645 & 1,270,830 \\
\quad Judge marks a name or code & 37,243 & 36,899 & 56,601 \\
\quad No candidates & 2,692 & 2,687 & 5,290 \\
\quad No kept synonym & 28,739 & 28,495 & 54,519 \\
\quad No single-token synonym & 36,598 & 36,478 & 51,664 \\
\quad Observed token $w_0$ in no concept & 46,632 & 46,254 & 119,816 \\
\quad Negating prefix & 6,294 & 6,275 & 7,735 \\
\quad Synonyms in no concept & 3,014 & 3,005 & 6,192 \\
\quad Observed token $w_0$ not in the closest concept $c^\star$ & 2,602 & 2,586 & 4,007 \\
\quad Lexical-support check & 6,606 & 6,591 & 7,738 \\
\quad Not a whole word in the document & 21,495 & 21,506 & 28,526 \\
\addlinespace[2pt]
\textbf{Concept-target positions} & \textbf{298,406} & \textbf{298,227} & \textbf{475,558} \\
\quad fraction of occurrences in kept windows (\%) & 14.3 & 14.3 & 22.8 \\
\bottomrule\end{tabular}
\caption{\textbf{From judged occurrences to concept targets.} Each check applies to the occurrences that passed the earlier ones. The rows follow the order of the checks, not the numbering of the 6 rules in ``Which occurrences get a target'' (Appendix~\ref{app:data-eligibility}). Two further checks, for a concept of 200 or more words and for a closest concept $c^\star$ with no member token, removed none. Occurrences outside the training documents' windows include those in the development and final held-out documents (164,408 / 164,411 / 164,458 for Llama / Qwen2.5 / Falcon3).}\label{tab:eligibility_funnel}
\end{table*}

\paragraph{Documents and windows.}
Training reads running text: \revnew{\emph{training windows}} of at most 256 tokens from the documents that hold collected occurrences. A document is tokenized whole and divided into consecutive windows. A document's final piece shorter than 32 tokens is dropped rather than trained as a nearly empty window, and so is a window that holds no collected occurrence. Windows carry no beginning-of-sequence token, so a window's first token is never predicted. The training files hold 814,799 windows for Llama, 820,136 for Qwen2.5 and 819,697 for Falcon3, and the 50,000 optimizer steps read the first 800,000 once, in one fixed random order of documents (each document's windows consecutive) shared by every method and seed.

Hashing each document's identifier holds out a random but fixed 10\% of the documents for the held-out evaluation of Section~\ref{sec:results}, half for development and half for final evaluation. The 17,326 documents of the concept-construction sample are never held out, so no held-out text helped build a concept. \revnew{A held-out occurrence gets a target as often as a training occurrence in documents outside the concept-construction sample (Llama 6.3\% against 6.4\%, Falcon3 12.2\% against 12.4\%).} Held-out files keep the content-word positions that get no target, for the content-word perplexity.

\paragraph{What the files hold.}
For Llama, the training windows hold 196.8 million predicted positions, and 32\% of them are content words (counted in every 400th window). Of these content words, 2,088,585 are collected occurrences (3.3\%), and 76.5\% of the collected occurrences have an observed word $w_0$ that splits into several tokens. The 298,406 that get concept targets are 0.15\% of all positions. Training targets have 7,875 distinct closest concepts $c^\star$ for Llama, 7,875 for Qwen2.5 and 8,464 for Falcon3, and held-out targets 2,523 for Llama, 2,523 for Qwen2.5 and 3,648 for Falcon3.

\paragraph{Known errors.}
\revnew{The lexical-support check (rule 6 of ``Which occurrences get a target'') removes some targets whose closest concept $c^\star$ is wrong. A correct removal: in Falcon3, the token of the musical \emph{viol} is the first token of \emph{violator} and belongs only to 2 offender concepts (\#2641 and \#5887), so its target, which went to \#2641, is removed. However, the lexical-support check can also remove useful targets and keep wrong senses: \emph{matt}, a dull finish, reaches a carpets concept (\#9016 \revnew{\{}\emph{carpets}, \emph{mats}, \emph{rugs}, \dots\revnew{\}}) through the first token of \emph{mattings}, and keeps its target in all 3 families because its synonym \emph{mat} has the same lemma as the member \emph{mats}. Wrong senses also pass where the check does not apply, because the observed word is itself a member word: in a \emph{light water} reactor, where \emph{light} means ordinary rather than heavy water, Falcon3's target for \emph{light} still goes to an illumination concept \revnew{(\#4930 \{\emph{illuminate}, \emph{light}, \emph{lit}, \dots\})}.} A misspelling can also become a member word: the judge kept \emph{safeguat}, whose first token makes \emph{safeg} the only token of \#1450 in just 2 concepts for Llama and Qwen2.5, and so sets \#1450's 50\% bound (Appendix~\ref{app:math}). These are examples, not an accuracy estimate.

\needspace{6\baselineskip}%
\section{Training and Evaluation \texorpdfstring{\rev{Settings}}{Settings}}
\label{app:configuration}
\revnew{Appendix~\ref{app:configuration} lists the post-training and evaluation settings in Table~\ref{tab:configuration}.} In every \rev{post-trained} method the retention loss $\mathcal L_{\mathrm{retention}}$ acts only at the other positions $\mathcal U$ (Equation~\ref{eq:retention}), so \rev{NTP post-training} trains the concept-target positions $\mathcal S$ with next-token prediction alone. \rev{On the same GPU in the same job, Conceptizer trains 5 to 8\% longer than NTP post-training for the 1B and 1.5B models and 1 to 4\% longer for the 3B to 8B models.}

\begin{table}[!htbp]\footnotesize\setlength{\tabcolsep}{3pt}
\noindent\begin{tabular}{@{}l@{\hspace{6pt}}p{0.68\linewidth}@{}}
\toprule
Setting & Value\\
\midrule
\multicolumn{2}{@{}l}{\emph{Post-training}}\\
Adapter & LoRA, rank 16, alpha 32 (updates scaled by alpha/rank${}=2$), dropout 0.05, no bias terms; the only trained parameters\\
Adapted layers & every linear attention and feed-forward projection (PEFT all-linear; \citealp{mangrulkar2022peft}); output head excluded\\
Precision & bfloat16 model weights\\
Hardware & 1 NVIDIA H100 per run; a 50,000-step run took about 7 to 9 hours of job time for the smallest models and about 20 hours for Llama~8B\\
Software & PyTorch 2.6.0 \citep{paszke2019pytorch}, Transformers 5.0.0 \citep{wolf2020transformers} and PEFT 0.19.1 for post-training and evaluation; NLTK 3.9.2 for the concept vocabulary\\
Optimizer & AdamW \citep{loshchilov2019adamw}, betas (0.9, 0.999), epsilon $10^{-8}$, weight decay 0.01; gradient norm clipped at 1.0\\
Learning rate & $7\times10^{-5}$\rev{, as in \citet{zhang2026concepts}}; linear warmup over 100 steps, then cosine decay \citep{loshchilov2017sgdr} to zero at step 50,000\\
Steps and batch & 50,000 optimizer steps of 16 windows of at most 256 tokens, no gradient accumulation; the final checkpoint is evaluated\\
Loss weights & $\lambda=0.6$ (concept), $\beta=0.5$ (retention); $\varepsilon=10^{-12}$ inside the concept logarithm\\
Training seeds & 0, 37 and 42 (Weighted NTP: seed 0, the smallest model of each family); a seed sets the LoRA initialization and dropout, and every seed reads the same windows in the same order\\
\addlinespace[4pt]
\multicolumn{2}{@{}l}{\emph{Knowledge and Commonsense}}\\
Harness & lm-evaluation-harness 0.4.10 \citep{gao2024harness}; each task's native prompt, zero-shot, no chat template; the answer with the highest likelihood is chosen, nothing is generated\\
Knowledge & test splits of ARC-Challenge (1,172 questions), ARC-Easy (2,376), OpenBookQA (500) and SciQ (1,000); acc\_norm; SciQ's prompt includes its supporting passage\\
Commonsense & validation splits of HellaSwag (10,042 questions; acc\_norm), COPA (100; acc) and WinoGrande (1,267; acc)\\
\addlinespace[4pt]
\multicolumn{2}{@{}l}{\emph{Sentiment and MTEB}}\\
Embeddings & MTEB 2.18.6; final-layer hidden states averaged over non-padding tokens in float32 (mean pooling, compared for decoder models by \citealp{behnamghader2024llm2vec}); inputs truncated at 256 tokens; batch 32; no instruction prompt; no L2 normalization\\
Sentiment & 9 classification tasks (\rev{Section~\ref{sec:results}}), the 6 outside the 40-task suite in MTEB's .v2 versions; mean test-split accuracy of MTEB's logistic-regression classifier over 10 draws of training examples (below)\\
MTEB & the 40 tasks of Appendix~\ref{app:mteb}; each task's recorded score\\
\addlinespace[4pt]
Group scores & equal weight for every task of a group\\
\bottomrule
\end{tabular}
\caption{\textbf{Post-training and evaluation settings.} One recipe for all 9 models and every post-trained method, with nothing tuned per model or size.}\label{tab:configuration}
\end{table}

\paragraph{COPA.}
COPA is scored on its 100 SuperGLUE validation questions \citep{wang2019superglue}, which weigh as much as the larger tasks.

\paragraph{The \rev{sentiment} classifier.}
MTEB's logistic-regression classifier (scikit-learn 1.8.0 \citep{pedregosa2011scikit} defaults, at most 100 iterations; MTEB 2.18.6 \texttt{abstasks/classification.py}) is trained on the embeddings of 8 to 128 training examples per label, MTEB's setting for each task (IMDb 8, ToxicConversations 16, TweetSentimentExtraction 32, Emotion 16, ToxicChat 8, PoemSentiment 8, FrenkEn 8, YelpReviewFull 128, AmazonPolarity 8).

\paragraph{The word-level comparison of Fig.~\ref{fig:word-example}.}
The figure reads \rev{the seed-0 Conceptizer and NTP post-training adapters of Llama 1B} at the running \rev{example's} training position with its full 256-token window: the probabilities of the single tokens of \emph{safety} (ID 7296), \emph{protection} (9313), \emph{prevention} (27344) and \emph{care} (2512), and of \emph{habits} (26870), \emph{related} (5552) and \emph{screen} (4264), the 3 largest absolute probability decreases among \rev{14,555} lowercase single-token member words of the concept vocabulary $\mathcal C$. Logits $\mathbf z_i$ are bfloat16 on CPU and the softmax is float64 over all 128,256 \rev{output tokens}, so \rev{tokens whose logits round to the same bfloat16 value get the same probability}.

\paragraph{Licenses and intended use.}
FineWeb-Edu is released under ODC-By 1.0 and WordNet 3.0 under Princeton's WordNet license. The post-trained models, the candidate generators and the judge are used under their own licenses: the Llama 3.1, 3.2 and 3.3 Community Licenses, Qwen's licenses for Qwen2.5, TII's Falcon license for Falcon3 and Apache 2.0 for gpt-oss-120B. The benchmarks are used for evaluation only, under their own terms. Our release, code under MIT and data derived from FineWeb-Edu, these models' outputs and WordNet, keeps the terms of its sources and is intended for research on language-model training.

\begin{table}[t]\centering\footnotesize\setlength{\tabcolsep}{2.2pt}\renewcommand{\arraystretch}{1}
\begin{tabular}[t]{@{}lrrrr@{}}
\toprule
Construction & Concepts $|\mathcal C|$ & Largest & $\geq$200 & \begin{tabular}[b]{@{}r@{}}Senses of\\\emph{light}\end{tabular} \\
\midrule
\multicolumn{5}{@{}l}{\emph{Dice threshold for senses}} \\
0.02 & 7,128 & 169 & 0 & 1 \\
0.05 & 7,611 & 169 & 0 & 1 \\
0.10 & 8,284 & 169 & 0 & 1 \\
0.15 (adopted) & 9,061 & 169 & 0 & 2 \\
0.20 & 9,878 & 192 & 0 & 2 \\
\addlinespace[3pt]
\multicolumn{5}{@{}l}{\emph{Jaccard threshold for concepts}} \\
0.25 & 6,679 & 223 & 1 & 1 \\
1/3 (adopted) & 7,611 & 169 & 0 & 1 \\
0.50 & 8,422 & 134 & 0 & 1 \\
\addlinespace[3pt]
\multicolumn{5}{@{}l}{\emph{Linkage for concepts}} \\
Single & 7,174 & 411 & 17 & 1 \\
Complete (adopted) & 7,611 & 169 & 0 & 1 \\
\bottomrule
\end{tabular}
\par\medskip
\begin{tabular}[t]{@{}lrrr@{}}
\toprule
 & \multicolumn{3}{c}{Held-out perplexity $\downarrow$} \\
\cmidrule(l){2-4}
Training & Concept-target & Content-word & All tokens \\
\midrule
\multicolumn{4}{@{}l}{\emph{Concept-loss weight $\lambda$}} \\
0.3 & 454.89 & 367.84 & 18.89 \\
1.0 & 356.71 & 373.40 & 19.71 \\
\addlinespace[3pt]
\multicolumn{4}{@{}l}{\emph{Randomized synonyms, scrambled draw}} \\
1 & 567.07 & 329.01 & 19.95 \\
2 & 563.20 & 328.21 & 19.94 \\
\bottomrule
\end{tabular}
\caption{\textbf{Concept-vocabulary counts under other construction settings and held-out perplexity under other training settings.} One setting changes per block. Top: the concept vocabulary $\mathcal C$ built from the concept-construction sample, where the adopted setting (Dice 0.15, Jaccard 1/3, complete linkage) gives the 9,061 concepts. The Dice block uses complete linkage at Jaccard 1/3, the Jaccard block Dice 0.05 with complete linkage, and the linkage block Dice 0.05 at Jaccard 1/3. Largest: words in the largest concept; $\geq$200: concepts with 200 or more words; Senses of \emph{light}: senses found for the verb \emph{light}. Bottom: Llama 1B after 50,000 optimizer steps (seed 0), at 5,428 concept-target positions, 82,014 content-word positions and 9,737,170 tokens (the scrambled draws at $\lambda=0.6$).}\label{tab:sensitivity}
\end{table}

\needspace{6\baselineskip}%
\section{Sensitivity to Construction and Training Settings}
\label{app:sensitivity}
\revnew{Appendix~\ref{app:sensitivity} shows how the concept vocabulary and the held-out perplexity change under other construction and training settings.}

\paragraph{\revnew{Construction settings.}}
\revnew{The construction sweeps rebuild the concept vocabulary from the concept-construction sample with one setting changed at a time (Table~\ref{tab:sensitivity}, top; Fig.~\ref{fig:construction-sweep}).} A higher Dice threshold splits words into more senses and so gives more concepts. The lowest tested value that gives the verb \emph{light} 2 senses is 0.15, although some of its pools still mix the ignite and illuminate \rev{meanings}. \revnew{A higher Jaccard threshold merges fewer senses and so gives more and smaller concepts. At Dice 0.05, the lowest tested value that leaves no concept of 200 or more words is 1/3. At 0.25, one concept has 223 words.} Single linkage chains senses into concepts of 200 or more words, and complete linkage does not.

\paragraph{\revnew{Training settings.}}
\revnew{The training sweeps train Llama 1B at 2 other concept-loss weights $\lambda$ and with 2 other scrambled draws of Randomized synonyms (Table~\ref{tab:sensitivity}, bottom). They are scored at 5,428 concept-target positions, not the 5,154 of Section~\ref{sec:heldout}, so they compare settings with each other.} A larger concept-loss weight $\lambda$ lowers concept-target perplexity and raises global perplexity, and the 2 scrambled draws give nearly equal values.
\ConstructionSweepFigure

\needspace{6\baselineskip}%
\section{Judged-Synonym Probability, Weighted NTP and Randomized Synonyms}
\label{app:controls}
\revnew{Appendix~\ref{app:controls} gives the details behind 3 results of Section~\ref{sec:results}: the judged-synonym probability, the alternative Weighted NTP and the content-word lead of Randomized synonyms.}

\paragraph{Judged-synonym probability.}
Every held-out concept-target position has at least one contextual synonym $w_j$ besides the observed word $w_0$, judged for that occurrence as for a training target, so the judged-synonym probability covers all of them. Table~\ref{tab:synonym-probability} and Fig.~\ref{fig:size-synonyms} report its arithmetic mean, which can be dominated by positions where the synonyms already hold much probability, and its geometric mean, which follows the typical position. Probabilities below $10^{-12}$ are raised to $10^{-12}$ before the logarithm, which happens only for Falcon3's base models, at 26 or fewer of their 10,028 positions. Randomized synonyms raises the geometric mean above NTP post-training's but lowers the arithmetic mean in every model.
\begin{table*}[t]\centering\footnotesize\setlength{\tabcolsep}{3pt}\renewcommand{\arraystretch}{1}
\begin{tabular}[t]{@{}lcccccccc@{}}
\toprule
 & \multicolumn{8}{c}{Judged-synonym probability (\%) $\uparrow$} \\
\cmidrule(l){2-9}
 & \multicolumn{2}{c}{} & \multicolumn{2}{c}{} & \multicolumn{2}{c}{Randomized} & \multicolumn{2}{c}{} \\
 & \multicolumn{2}{c}{Base model} & \multicolumn{2}{c}{NTP post-training} & \multicolumn{2}{c}{synonyms} & \multicolumn{2}{c}{Conceptizer} \\
\cmidrule(lr){2-3}\cmidrule(lr){4-5}\cmidrule(lr){6-7}\cmidrule(lr){8-9}
Model & Geo. & Arith. & Geo. & Arith. & Geo. & Arith. & Geo. & Arith. \\
\midrule
Llama 1B & 0.192 & 3.94 & 0.216 & \underline{3.96} & \underline{0.313} & 2.59 & \textbf{0.464} & \textbf{5.00} \\
Llama 3B & 0.223 & \underline{4.63} & 0.251 & 4.62 & \underline{0.368} & 2.90 & \textbf{0.619} & \textbf{6.21} \\
Llama 8B & 0.242 & \underline{5.33} & 0.274 & 5.18 & \underline{0.419} & 3.24 & \textbf{0.774} & \textbf{7.38} \\
Qwen2.5 1.5B & 0.223 & \underline{4.60} & 0.239 & 4.53 & \underline{0.378} & 2.81 & \textbf{0.582} & \textbf{5.86} \\
Qwen2.5 3B & 0.220 & \underline{5.10} & 0.254 & 4.97 & \underline{0.410} & 3.04 & \textbf{0.672} & \textbf{6.50} \\
Qwen2.5 7B & 0.207 & \underline{5.51} & 0.261 & 5.39 & \underline{0.434} & 3.35 & \textbf{0.765} & \textbf{7.31} \\
Falcon3 1B & 0.187 & \underline{4.78} & 0.231 & 4.47 & \underline{0.344} & 2.44 & \textbf{0.580} & \textbf{5.50} \\
Falcon3 3B & 0.166 & \underline{5.18} & 0.237 & 4.69 & \underline{0.361} & 2.52 & \textbf{0.651} & \textbf{5.91} \\
Falcon3 7B & 0.153 & \underline{5.99} & 0.278 & 5.47 & \underline{0.433} & 2.93 & \textbf{0.862} & \textbf{7.16} \\
\bottomrule
\end{tabular}
\caption{\textbf{Conceptizer has the highest judged-synonym probability against the 3 baselines in every model, by both means.} At held-out concept-target positions at seed 0, across all model sizes. Geo., Arith.: geometric and arithmetic means over positions.}\label{tab:synonym-probability}
\end{table*}

\SizeSynonymFigure

\paragraph{Weighted NTP.}
Weighted NTP keeps the NTP and retention losses, $\mathcal L_{\mathrm{NTP}}$ and $\mathcal L_{\mathrm{retention}}$, and replaces the concept loss $\mathcal L_{\mathrm{concept}}$ with the mean cross-entropy of the observed word $w_0$ over the same concept-target positions $\mathcal S$, at the same weight $\lambda=0.6$. In an optimizer step, the observed word $w_0$ at a concept-target position thus gets weight $1/|\mathcal I|+\lambda/|\mathcal S|$ and every other position $1/|\mathcal I|$. Everything else is shared. The name follows the Weighted NTP control of \citet{zhang2026concepts}, which gives content words a fixed weight (they report weight 5, whose global perplexity is closest to that of their concept-trained models). Ours reuses the concept loss's weight $\lambda$ and averaging instead, which gives a concept-target position about 400 times the weight of another position in an average Llama optimizer step (5.9 targets among 3,865 positions).

Table~\ref{tab:weighted-ntp} compares it with Conceptizer. Weighted NTP has the lowest concept-target perplexity of all post-trained methods in all 3 models at the smallest size. Its knowledge, sentiment and commonsense scores are comparable to Conceptizer's. However, it has a lower judged-synonym probability and a higher global perplexity than Conceptizer in all 3.
\begin{table}[t]\centering\footnotesize\setlength{\tabcolsep}{2pt}\renewcommand{\arraystretch}{1}
\begin{tabular}[t]{@{}lccc@{}}
\toprule
 & \multicolumn{3}{c}{Held-out perplexity $\downarrow$} \\
\cmidrule(l){2-4}
Method & Concept-target & Content-word & All tokens \\
\midrule
\multicolumn{4}{@{}l}{\textbf{Llama 1B}} \\
Weighted NTP & \textbf{218.54} & 376.12 & 19.66 \\
Conceptizer & 367.16 & \textbf{373.71} & \textbf{19.31} \\
\addlinespace[3pt]
\multicolumn{4}{@{}l}{\textbf{Qwen2.5 1.5B}} \\
Weighted NTP & \textbf{142.95} & 235.43 & 14.73 \\
Conceptizer & 231.71 & \textbf{232.09} & \textbf{14.48} \\
\addlinespace[3pt]
\multicolumn{4}{@{}l}{\textbf{Falcon3 1B}} \\
Weighted NTP & \textbf{184.93} & \textbf{403.42} & 18.25 \\
Conceptizer & 306.40 & 410.92 & \textbf{17.38} \\
\bottomrule
\end{tabular}
\par\medskip
\begin{tabular}[t]{@{}lcccc@{}}
\toprule
 & \multicolumn{4}{c}{Percent $\uparrow$} \\
\cmidrule(l){2-5}
Method & \begin{tabular}[b]{@{}c@{}}Judged-\\synonym\end{tabular} & Knowledge & Sentiment & \begin{tabular}[b]{@{}c@{}}Common-\\sense\end{tabular} \\
\midrule
\multicolumn{5}{@{}l}{\textbf{Llama 1B}} \\
Weighted NTP & 0.298 & \textbf{59.45} & 56.97 & 68.52 \\
Conceptizer & \textbf{0.464} & 59.22 & \textbf{57.13} & \textbf{68.71} \\
\addlinespace[3pt]
\multicolumn{5}{@{}l}{\textbf{Qwen2.5 1.5B}} \\
Weighted NTP & 0.330 & 64.76 & \textbf{55.57} & \textbf{72.95} \\
Conceptizer & \textbf{0.582} & \textbf{65.36} & 55.48 & 72.31 \\
\addlinespace[3pt]
\multicolumn{5}{@{}l}{\textbf{Falcon3 1B}} \\
Weighted NTP & 0.392 & \textbf{64.39} & 57.50 & \textbf{66.35} \\
Conceptizer & \textbf{0.580} & 63.95 & \textbf{57.54} & 66.08 \\
\bottomrule
\end{tabular}
\caption{\textbf{Weighted NTP has lower concept-target perplexity than Conceptizer in all 3 models, but has a lower judged-synonym probability and a higher global perplexity.} Seed 0 at the smallest model of each family. Judged-synonym: the geometric mean of Table~\ref{tab:synonym-probability}; Knowledge, Sentiment and Commonsense: the mean of each group's tasks.}\label{tab:weighted-ntp}
\end{table}

\paragraph{Randomized synonyms on content words.}
Most content-word positions hold no concept target: for Llama, 76,860 of 82,014 do not. Of these, 68,746 are the first token of a word that splits into several tokens, such as \emph{embed} in \emph{embeddable}. Randomized synonyms' lead comes entirely from these positions: in all 9 models at seed 0, on first tokens that belong to concepts through first-token memberships it gives 1.1 to 1.4 times Conceptizer's probability, while on words that are a single token Conceptizer is higher. Scored over the whole word, the lead disappears: on 3,468 sampled multi-token words of Llama 1B's held-out text, Randomized synonyms gives its first-token gain back on the remaining tokens and matches NTP post-training (7.79 against 7.79 nats per word), while Conceptizer is 0.17 nats above both.

Our intuition is that scrambled synonyms send the reward of the concept loss $\mathcal L_{\mathrm{concept}}$ to concepts unrelated to the sentence, and first-token memberships pass it on to the tokens that begin many member words: for Llama, \emph{un-} and \emph{re-} belong to 447 and 471 concepts, against 7 for the median member token. On sampled training batches of Llama 1B, the scrambled targets put about 3 times as much upward push on the tokens that belong to concepts only through first-token memberships (14 to 16\% of the upward push on all logits, against 4 to 6\% for Conceptizer's targets). Scrambling the synonyms alone reproduces this, and redrawing the closest concept $c^\star$ alone does not. These are gradients measured at the trained models, not along training. We expect the lead to disappear once concept targets also cover words that split into several tokens.

\needspace{6\baselineskip}%
\section{MTEB}
\label{app:mteb}
\revnew{Appendix~\ref{app:mteb} gives the results on the 40-task MTEB suite.} The MTEB suite is MTEB's English benchmark, version 2 \citep{muennighoff2023mteb,enevoldsen2025mmteb}, without MindSmallReranking \citep{wu2020mind}, whose data did not fit in the evaluation job's memory, so it was scored for no model. Its 40 tasks fall in 7 sections: 8 classification, 8 clustering, 3 pair classification, 1 reranking, 10 retrieval, 9 semantic textual similarity and 1 summarization, where SummEval \citep{fabbri2021summeval} scores supplied summaries and nothing is generated. Every task uses the embeddings of Table~\ref{tab:configuration} and its own MTEB metric. MTEB's recorded score averages the validation and test splits for 4 classification tasks (AmazonCounterfactual \citep{oneill2021counterfactual}, MTOPDomain \citep{li2021mtop}, MassiveIntent and MassiveScenario \citep{bastianelli2020slurp,fitzgerald2023massive}) and is the test score for the other 36. Table~\ref{tab:mteb} and Fig.~\ref{fig:mteb} give the results, and 3 of the suite's classification tasks (IMDb, ToxicConversations and TweetSentimentExtraction) also belong to the \rev{sentiment} group. The other 32 tasks come from these sources: in classification, Banking77 \citep{casanueva2020banking77}; in clustering, StackExchange \citep{geigle2021tweac}, TwentyNewsgroups \citep{lang1995newsweeder}, and ArXivHierarchical (P2P and S2S), BiorxivP2P, MedrxivP2P, MedrxivS2S and StackExchangeP2P, which MTEB built itself; in pair classification, SprintDuplicateQuestions \citep{shah2018sprint}, TwitterSemEval2015 \citep{xu2014twitter,xu2015pit} and TwitterURLCorpus \citep{lan2017twitterurl}; in reranking, AskUbuntuDupQuestions \citep{lei2016askubuntu}; in retrieval, BEIR's versions \citep{thakur2021beir} of ArguAna \citep{wachsmuth2018arguana}, CQADupstackGaming and CQADupstackUnix \citep{hoogeveen2015cqadupstack}, ClimateFEVER \citep{diggelmann2020climatefever}, FEVER \citep{thorne2018fever}, FiQA2018 \citep{maia2018fiqa}, HotpotQA \citep{yang2018hotpotqa}, SCIDOCS \citep{cohan2020specter}, TRECCOVID \citep{wang2020cord19,voorhees2020treccovid} and Touche2020 \citep{ajjour2019argsme,bondarenko2020touche}, with MTEB's hard-negative versions of ClimateFEVER, FEVER and HotpotQA and version 3 of Touche2020 \citep{thakur2024touche}; and in semantic textual similarity, BIOSSES \citep{sogancioglu2017biosses}, SICK-R \citep{marelli2014sick}, STS12 to STS15 \citep{agirre2012sts,agirre2013sts,agirre2014sts,agirre2015sts}, STS17 and STSBenchmark \citep{cer2017sts} and STS22 \citep{chen2022sts}.
\MtebFigure
\begin{table*}[tp]\centering\footnotesize\setlength{\tabcolsep}{3pt}\renewcommand{\arraystretch}{1}
\begin{tabular}[t]{@{}llrrrrrrrr@{}}
\toprule
 & & \multicolumn{8}{c}{Recorded MTEB score $\times$100 (number of tasks) $\uparrow$} \\
\cmidrule(l){3-10}
Model & Method & All (40) & Class. (8) & Clust. (8) & Pair (3) & Rerank. (1) & Retr. (10) & STS (9) & Summ. (1) \\
\midrule
Llama 1B & Base model & \textbf{43.02} & \textbf{67.58} & \underline{41.76} & \textbf{56.43} & \textbf{15.49} & \textbf{13.92} & \textbf{56.40} & 14.48 \\
 & NTP post-training & \underline{42.41} & 66.54 & \textbf{41.98} & \underline{54.88} & 14.96 & \underline{13.63} & \underline{55.46} & 13.26 \\
 & Rand. synonyms & 41.63 & 66.49 & 41.36 & 52.80 & 14.51 & 12.91 & 53.94 & \underline{14.85} \\
 & Conceptizer & 42.33 & \underline{67.02} & 41.74 & 54.25 & \underline{15.47} & 13.33 & 55.11 & \textbf{15.45} \\
\addlinespace[2pt]
Llama 3B & Base model & 42.62 & \textbf{67.79} & 41.72 & \textbf{58.51} & \textbf{16.23} & 12.63 & \textbf{56.16} & 5.21 \\
 & NTP post-training & \textbf{42.73} & 66.85 & \textbf{42.48} & \underline{57.29} & \underline{15.51} & \textbf{14.15} & \underline{55.43} & 6.74 \\
 & Rand. synonyms & 41.74 & 66.76 & 41.79 & 54.82 & 15.15 & 12.74 & 53.87 & \underline{9.26} \\
 & Conceptizer & \underline{42.65} & \underline{67.34} & \underline{42.17} & 56.30 & 15.48 & \underline{13.82} & 55.29 & \textbf{9.82} \\
\addlinespace[2pt]
Llama 8B & Base model & 41.87 & \textbf{66.35} & \underline{43.56} & \textbf{58.13} & \textbf{14.88} & 11.42 & 54.21 & 4.10 \\
 & NTP post-training & \underline{42.24} & 66.05 & \textbf{43.87} & \underline{56.98} & 14.55 & \underline{12.94} & \textbf{54.43} & 5.41 \\
 & Rand. synonyms & 41.10 & 65.50 & 43.19 & 53.32 & 13.53 & 11.79 & 52.56 & \underline{9.92} \\
 & Conceptizer & \textbf{42.40} & \underline{66.21} & \underline{43.56} & 56.41 & \underline{14.72} & \textbf{13.39} & \underline{54.39} & \textbf{10.50} \\
\addlinespace[2pt]
Qwen2.5 1.5B & Base model & 36.91 & \textbf{62.58} & \textbf{40.42} & 52.51 & \underline{9.86} & \underline{8.71} & 44.29 & \textbf{{\boldmath$-$}0.91} \\
 & NTP post-training & \textbf{37.89} & 62.54 & \underline{40.37} & \textbf{53.58} & \textbf{10.60} & 8.69 & \textbf{48.58} & $-$3.21 \\
 & Rand. synonyms & 36.73 & 61.79 & 39.48 & 50.31 & 9.11 & 8.17 & 46.51 & \underline{$-$1.17} \\
 & Conceptizer & \underline{37.82} & \underline{62.56} & 40.23 & \underline{52.60} & 9.69 & \textbf{8.83} & \underline{48.48} & $-$1.50 \\
\addlinespace[2pt]
Qwen2.5 3B & Base model & 37.13 & 63.42 & \underline{40.46} & 52.46 & \textbf{10.98} & 8.94 & 44.06 & \underline{$-$0.10} \\
 & NTP post-training & \textbf{38.38} & \underline{63.75} & \textbf{40.56} & \textbf{54.10} & \underline{10.69} & \textbf{9.20} & \textbf{48.48} & $-$0.74 \\
 & Rand. synonyms & 36.73 & 62.98 & 39.56 & 50.51 & 9.82 & 8.05 & 45.24 & $-$0.28 \\
 & Conceptizer & \underline{38.24} & \textbf{63.79} & 40.40 & \underline{52.81} & 10.27 & \underline{9.17} & \underline{48.41} & \textbf{{\boldmath$-$}0.03} \\
\addlinespace[2pt]
Qwen2.5 7B & Base model & 37.63 & \textbf{62.82} & \underline{41.51} & \underline{56.61} & \underline{10.13} & \textbf{8.53} & 44.90 & 1.23 \\
 & NTP post-training & \textbf{38.09} & \underline{62.08} & \textbf{41.56} & \textbf{56.88} & \textbf{10.57} & \underline{8.43} & \underline{47.77} & $-$1.05 \\
 & Rand. synonyms & 36.26 & 60.70 & 40.35 & 53.39 & 8.79 & 6.67 & 44.96 & \textbf{1.59} \\
 & Conceptizer & \underline{37.92} & 61.66 & 41.41 & 56.14 & 9.75 & 8.08 & \textbf{47.97} & \underline{1.49} \\
\addlinespace[2pt]
Falcon3 1B & Base model & 40.31 & 66.31 & 41.10 & 50.95 & 10.74 & 12.39 & 50.70 & 9.19 \\
 & NTP post-training & \underline{42.29} & \underline{66.84} & \textbf{42.04} & \textbf{54.40} & \underline{11.90} & 13.93 & \underline{55.11} & 10.02 \\
 & Rand. synonyms & 42.01 & 66.76 & 41.44 & 51.53 & 11.74 & \underline{14.23} & 54.73 & \textbf{13.74} \\
 & Conceptizer & \textbf{42.60} & \textbf{67.11} & \underline{41.87} & \underline{53.59} & \textbf{11.98} & \textbf{14.33} & \textbf{55.92} & \underline{12.83} \\
\addlinespace[2pt]
Falcon3 3B & Base model & 40.81 & \underline{66.60} & 41.90 & 54.33 & 10.30 & 11.84 & 52.58 & $-$0.62 \\
 & NTP post-training & \underline{41.91} & \underline{66.60} & \textbf{42.41} & \textbf{56.07} & \underline{11.40} & 12.63 & \underline{54.96} & \underline{3.86} \\
 & Rand. synonyms & 41.40 & 66.33 & 41.55 & 52.67 & 10.94 & \underline{13.28} & 54.17 & 3.56 \\
 & Conceptizer & \textbf{42.28} & \textbf{66.81} & \underline{42.24} & \underline{54.70} & \textbf{11.71} & \textbf{13.59} & \textbf{55.65} & \textbf{6.04} \\
\addlinespace[2pt]
Falcon3 7B & Base model & 40.79 & 65.15 & 41.29 & 54.64 & \underline{11.21} & 12.74 & 52.11 & 8.57 \\
 & NTP post-training & \textbf{41.78} & \underline{65.46} & \textbf{42.07} & \textbf{56.54} & \textbf{11.76} & \textbf{13.44} & \textbf{54.17} & 7.82 \\
 & Rand. synonyms & 40.24 & 64.76 & 40.80 & 52.93 & 10.36 & 12.29 & 51.59 & \underline{8.72} \\
 & Conceptizer & \underline{41.35} & \textbf{65.49} & \underline{41.50} & \underline{55.45} & 10.86 & \underline{13.37} & \underline{53.10} & \textbf{9.40} \\
\bottomrule
\end{tabular}
\caption{\textbf{On the whole 40-task MTEB suite, results are mixed: the best mean belongs to NTP post-training in 5 models, Conceptizer in 3 and the base model in 1.} Equal weights over the 40 tasks and within each section; means of training seeds 0, 37 and 42. Rand. synonyms: Randomized synonyms.}\label{tab:mteb}
\end{table*}

\needspace{6\baselineskip}%
\section{Prompts}
\label{app:prompts}
\revnew{Appendix~\ref{app:prompts} gives the prompts of the 3 LLMs: Llama-3.3-70B-Instruct and Qwen2.5-72B-Instruct, which propose synonym candidates, and gpt-oss-120B, the judge.} Each model receives a system message, worked examples as earlier user and assistant turns, and then the input, whose slots are shown in italics. The texts are exact.

\needspace{5\baselineskip}%
\medskip
\noindent\textbf{Llama-3.3-70B-Instruct, candidates in context.}\par\smallskip
{\footnotesize\ttfamily\raggedright\linespread{1.1}\selectfont\setlength{\parskip}{3pt plus 1pt}
{\rmfamily\itshape System:}\par
You are an English lexical-substitution annotator producing CANDIDATE synonyms. You are given a SENTENCE with one TARGET word marked in «double angle brackets» and its part of speech. List the words that could replace the TARGET *in this sentence* with the same meaning.\par
This is a high-RECALL step — a separate precision filter runs afterward — so cast a WIDE net: include both close synonyms and looser near-synonyms, as long as each plausibly preserves the meaning in this context.\par
Rules:\\
- Same part of speech and grammatical form as the target (it must drop in directly).\\
- SINGLE WORDS ONLY — no multi-word phrases, and never join words together: if a fitting replacement cannot be written as one ordinary word, omit it (do not output things like "celldeath" or "alkalineearthmetal").\\
- Never the target word itself; no duplicates; no direct antonyms.\\
- Err toward MORE candidates; rank closest-meaning first; up to 15.\\
- Output [] only if genuinely nothing could replace it.\\
- Output ONLY a JSON array of lowercase strings, nothing else.\par\medskip
{\rmfamily\itshape User:}\par
Sentence: The waiter carefully «placed» the warm plate on the table.\\
Target: "placed" (verb)\par
{\rmfamily\itshape Assistant:}\par
{}["set", "put", "laid", "positioned", "rested", "deposited", "planted"]\par\medskip
{\rmfamily\itshape User:}\par
Sentence: The committee will «convene» next Tuesday to vote.\\
Target: "convene" (verb)\par
{\rmfamily\itshape Assistant:}\par
{}["meet", "gather", "assemble", "congregate", "convoke"]\par\medskip
{\rmfamily\itshape User:}\par
Sentence: Marie Curie isolated «radium» in 1898.\\
Target: "radium" (noun)\par
{\rmfamily\itshape Assistant:}\par
{}[]\par\medskip
{\rmfamily\itshape Input:}\par
Sentence: {\rmfamily\itshape the sentence, with the word marked}\\
Target: "{\rmfamily\itshape \rev{word as written}}" ({\rmfamily\itshape noun, verb or adjective})\par}

\needspace{5\baselineskip}%
\bigskip
\noindent\textbf{Qwen2.5-72B-Instruct, candidates without context.}\par\smallskip
{\footnotesize\ttfamily\raggedright\linespread{1.1}\selectfont\setlength{\parskip}{3pt plus 1pt}
{\rmfamily\itshape System:}\par
You are an English thesaurus. Given a WORD and its part of speech, list the single-word synonyms that could replace it in SOME context, across ALL of its meanings.\par
This is a comprehensive, context-free step: cover EVERY sense the word has, and include both close and looser near-synonyms. A later step selects the right ones for each context.\par
Rules:\\
- Same part of speech as the word.\\
- SINGLE WORDS ONLY — no multi-word phrases, and never join words together.\\
- Never the word itself; no duplicates; no direct antonyms.\\
- Cover ALL senses: if the word has several meanings, include synonyms of each; up to 25.\\
- Output [] only if the word has no synonyms in any sense (e.g. a proper-noun name or opaque code).\\
- Output ONLY a JSON array of lowercase strings, nothing else.\par\medskip
{\rmfamily\itshape User:}\par
Word: "convene" (verb)\par
{\rmfamily\itshape Assistant:}\par
{}["meet", "gather", "assemble", "congregate", "convoke", "summon"]\par\medskip
{\rmfamily\itshape User:}\par
Word: "bank" (noun)\par
{\rmfamily\itshape Assistant:}\par
{}["shore", "edge", "slope", "embankment", "treasury", "lender", "reserve", "row", "tier", "heap", "mound"]\par\medskip
{\rmfamily\itshape User:}\par
Word: "radium" (noun)\par
{\rmfamily\itshape Assistant:}\par
{}[]\par\medskip
{\rmfamily\itshape Input:}\par
Word: "{\rmfamily\itshape lemma}" ({\rmfamily\itshape noun, verb or adjective})\par}

\needspace{5\baselineskip}%
\bigskip
\noindent\textbf{gpt-oss-120B, the LLM judge.}\par\smallskip
{\footnotesize\ttfamily\raggedright\linespread{1.1}\selectfont\setlength{\parskip}{3pt plus 1pt}
{\rmfamily\itshape System:}\par
You are a precise English lexical-substitution judge. You are given a SENTENCE with one TARGET word marked in «double angle brackets», its part of speech, and a LIST of candidate replacement words. Do two things.\par
STEP 1 — QUALITY CHECK. Decide whether the TARGET is even a concept word:\\
- If it is a PROPER NOUN (a specific person, place, organization, or product name) or an opaque code/abbreviation, it is NOT a concept -> set "drop": true and "synonyms": [].\\
- A genuinely technical CONCEPT (e.g. radium, apoptosis, covalent, hypothalamus) is a real concept -> "drop": false, even if it has no good synonyms.\par
STEP 2 — DISCRIMINATE (only if drop is false). Keep only the candidates that could replace the TARGET *in this sentence* without changing its meaning — true in-context synonyms of the target's specific sense here.\\
- Same part of speech and grammatical form as the target.\\
- SINGLE WORDS ONLY — no multi-word phrases; never the target word itself or any phrase containing it.\\
- Drop wrong-sense words, antonyms, broader/narrower terms, and the target itself.\\
- You may ADD up to 2 obvious in-context synonyms missing from the list.\\
- If nothing fits, "synonyms": [] (this is fine — a real concept can have no synonyms here).\par
Output ONLY a JSON object: \{"drop": <true|false>, "synonyms": [<lowercase words>]\}. No other text.\par\medskip
{\rmfamily\itshape User:}\par
Sentence: She carefully «placed» the book on the shelf.\\
Target: "placed" (verb)\\
Candidates: [set, put, laid, dropped, removed]\par
{\rmfamily\itshape Assistant:}\par
\{"drop": false, "synonyms": ["set", "put", "laid"]\}\par\medskip
{\rmfamily\itshape User:}\par
Sentence: The «bank» was muddy after the heavy flood.\\
Target: "bank" (noun)\\
Candidates: [shore, edge, slope, treasury, lender]\par
{\rmfamily\itshape Assistant:}\par
\{"drop": false, "synonyms": ["shore", "edge", "slope"]\}\par\medskip
{\rmfamily\itshape User:}\par
Sentence: The mayor met «Diana» at the new library opening.\\
Target: "Diana" (noun)\\
Candidates: [woman, girl, name, leader]\par
{\rmfamily\itshape Assistant:}\par
\{"drop": true, "synonyms": []\}\par\medskip
{\rmfamily\itshape User:}\par
Sentence: Marie Curie isolated «radium» in 1898.\\
Target: "radium" (noun)\\
Candidates: [element, metal, substance, isotope]\par
{\rmfamily\itshape Assistant:}\par
\{"drop": false, "synonyms": []\}\par\medskip
{\rmfamily\itshape Input:}\par
Sentence: {\rmfamily\itshape the sentence, with the word marked}\\
Target: "{\rmfamily\itshape \rev{word as written}}" ({\rmfamily\itshape noun, verb or adjective})\\
Candidates: [{\rmfamily\itshape all candidates of the \rev{3} sources, in alphabetical order, separated by commas}]\par}

\needspace{6\baselineskip}%
\section{Results for Every \texorpdfstring{\rev{Training Seed}}{Training Seed}}
\label{app:seeds}
\revnew{Appendix~\ref{app:seeds} gives every number behind the 3-seed means of Section~\ref{sec:results}.} Tables~\ref{tab:seed-knowledge}, \ref{tab:seed-sentiment}, \ref{tab:seed-commonsense} and~\ref{tab:seed-perplexity} give every run's scores, and Tables~\ref{tab:tasks-knowledge}, \ref{tab:tasks-commonsense} and~\ref{tab:tasks-sentiment} every task's 3-seed mean. Fig.~\ref{fig:size-accuracy} plots accuracy against model size, and Fig.~\ref{fig:training-curves} development-set perplexity during training.

Intervals are pointwise 95\% Student-$t$ intervals over the 3 training seeds (2 degrees of freedom), on fixed evaluation items. \revnew{Each evaluation job scores the post-trained methods of a seed together on one H100 GPU.} The paired intervals of Section~\ref{sec:results} come from these per-seed scores: differences are paired within seed, and a difference from the base model subtracts its single run from each seed's score. All Randomized synonyms runs share a single scrambled draw (2 other draws in Appendix~\ref{app:sensitivity}).
\begin{table}[tp]\centering\footnotesize\setlength{\tabcolsep}{3pt}\renewcommand{\arraystretch}{1}
\begin{tabular}[t]{@{}lrrrrr@{}}
\toprule
 & \multicolumn{5}{c}{Accuracy (\%) $\uparrow$} \\
\cmidrule(l){2-6}
Method & Seed 0 & 37 & 42 & Mean & 95\% CI \\
\midrule
\multicolumn{6}{@{}l}{\textbf{Llama 1B}} \\
Base model & \multicolumn{3}{c}{one fixed run} & 57.79 &  \\
NTP post-training & 58.01 & 58.49 & 57.99 & 58.16 & $\pm$0.70 \\
Rand. synonyms & 58.62 & 58.89 & 58.90 & \underline{58.80} & $\pm$0.40 \\
Conceptizer & 59.22 & 59.33 & 59.57 & \textbf{59.38} & $\pm$0.44 \\
\addlinespace[3pt]
\multicolumn{6}{@{}l}{\textbf{Llama 3B}} \\
Base model & \multicolumn{3}{c}{one fixed run} & 62.44 &  \\
NTP post-training & 62.96 & 63.29 & 63.07 & 63.11 & $\pm$0.42 \\
Rand. synonyms & 63.21 & 63.52 & 62.75 & \underline{63.16} & $\pm$0.97 \\
Conceptizer & 64.04 & 63.96 & 63.86 & \textbf{63.95} & $\pm$0.23 \\
\addlinespace[3pt]
\multicolumn{6}{@{}l}{\textbf{Llama 8B}} \\
Base model & \multicolumn{3}{c}{one fixed run} & \textbf{68.96} &  \\
NTP post-training & 67.76 & 67.98 & 68.34 & 68.03 & $\pm$0.73 \\
Rand. synonyms & 67.62 & 67.67 & 68.11 & 67.80 & $\pm$0.67 \\
Conceptizer & 68.98 & 67.87 & 68.85 & \underline{68.57} & $\pm$1.51 \\
\addlinespace[3pt]
\multicolumn{6}{@{}l}{\textbf{Qwen2.5 1.5B}} \\
Base model & \multicolumn{3}{c}{one fixed run} & 64.16 &  \\
NTP post-training & 64.01 & 64.05 & 63.82 & 63.96 & $\pm$0.31 \\
Rand. synonyms & 64.48 & 64.13 & 64.25 & \underline{64.28} & $\pm$0.44 \\
Conceptizer & 65.36 & 65.17 & 65.06 & \textbf{65.20} & $\pm$0.38 \\
\addlinespace[3pt]
\multicolumn{6}{@{}l}{\textbf{Qwen2.5 3B}} \\
Base model & \multicolumn{3}{c}{one fixed run} & 63.44 &  \\
NTP post-training & 65.63 & 65.56 & 66.19 & 65.80 & $\pm$0.86 \\
Rand. synonyms & 66.96 & 66.51 & 66.62 & \underline{66.70} & $\pm$0.58 \\
Conceptizer & 67.51 & 67.69 & 67.89 & \textbf{67.69} & $\pm$0.47 \\
\addlinespace[3pt]
\multicolumn{6}{@{}l}{\textbf{Qwen2.5 7B}} \\
Base model & \multicolumn{3}{c}{one fixed run} & 69.66 &  \\
NTP post-training & 70.54 & 70.81 & 70.98 & 70.78 & $\pm$0.55 \\
Rand. synonyms & 70.93 & 70.82 & 71.18 & \underline{70.98} & $\pm$0.47 \\
Conceptizer & 71.44 & 71.55 & 71.53 & \textbf{71.51} & $\pm$0.14 \\
\addlinespace[3pt]
\multicolumn{6}{@{}l}{\textbf{Falcon3 1B}} \\
Base model & \multicolumn{3}{c}{one fixed run} & 61.76 &  \\
NTP post-training & 62.26 & 62.17 & 62.39 & 62.27 & $\pm$0.28 \\
Rand. synonyms & 62.81 & 62.69 & 62.89 & \underline{62.80} & $\pm$0.25 \\
Conceptizer & 63.95 & 64.05 & 64.20 & \textbf{64.06} & $\pm$0.31 \\
\addlinespace[3pt]
\multicolumn{6}{@{}l}{\textbf{Falcon3 3B}} \\
Base model & \multicolumn{3}{c}{one fixed run} & \underline{66.73} &  \\
NTP post-training & 65.60 & 65.78 & 65.83 & 65.74 & $\pm$0.31 \\
Rand. synonyms & 65.42 & 65.64 & 65.44 & 65.50 & $\pm$0.31 \\
Conceptizer & 66.88 & 66.89 & 66.63 & \textbf{66.80} & $\pm$0.36 \\
\addlinespace[3pt]
\multicolumn{6}{@{}l}{\textbf{Falcon3 7B}} \\
Base model & \multicolumn{3}{c}{one fixed run} & \underline{69.50} &  \\
NTP post-training & 69.42 & 69.25 & 69.23 & 69.30 & $\pm$0.26 \\
Rand. synonyms & 69.10 & 69.25 & 69.08 & 69.14 & $\pm$0.24 \\
Conceptizer & 70.28 & 70.32 & 70.57 & \textbf{70.39} & $\pm$0.39 \\
\bottomrule
\end{tabular}
\caption{\textbf{Knowledge group at every training seed: Conceptizer has the best mean against the 3 baselines in 8 of 9 models.} The mean of ARC-Challenge, ARC-Easy, OpenBookQA and SciQ; 95\% CI: margin of error of the Student-$t$ training-seed interval. Rand. synonyms: Randomized synonyms.}\label{tab:seed-knowledge}
\end{table}

\begin{table}[tp]\centering\footnotesize\setlength{\tabcolsep}{3pt}\renewcommand{\arraystretch}{1}
\begin{tabular}[t]{@{}lrrrrr@{}}
\toprule
 & \multicolumn{5}{c}{Test-split accuracy (\%) $\uparrow$} \\
\cmidrule(l){2-6}
Method & Seed 0 & 37 & 42 & Mean & 95\% CI \\
\midrule
\multicolumn{6}{@{}l}{\textbf{Llama 1B}} \\
Base model & \multicolumn{3}{c}{one fixed run} & \underline{56.89} &  \\
NTP post-training & 56.19 & 56.37 & 56.38 & 56.31 & $\pm$0.26 \\
Rand. synonyms & 55.90 & 56.17 & 55.96 & 56.01 & $\pm$0.36 \\
Conceptizer & 57.13 & 57.20 & 57.27 & \textbf{57.20} & $\pm$0.17 \\
\addlinespace[3pt]
\multicolumn{6}{@{}l}{\textbf{Llama 3B}} \\
Base model & \multicolumn{3}{c}{one fixed run} & \underline{56.96} &  \\
NTP post-training & 56.73 & 56.68 & 56.66 & 56.69 & $\pm$0.09 \\
Rand. synonyms & 56.30 & 56.39 & 56.32 & 56.34 & $\pm$0.12 \\
Conceptizer & 57.60 & 57.79 & 57.70 & \textbf{57.70} & $\pm$0.24 \\
\addlinespace[3pt]
\multicolumn{6}{@{}l}{\textbf{Llama 8B}} \\
Base model & \multicolumn{3}{c}{one fixed run} & \underline{56.95} &  \\
NTP post-training & 56.98 & 56.94 & 56.84 & 56.92 & $\pm$0.18 \\
Rand. synonyms & 56.06 & 56.44 & 56.34 & 56.28 & $\pm$0.49 \\
Conceptizer & 57.72 & 57.82 & 57.75 & \textbf{57.76} & $\pm$0.13 \\
\addlinespace[3pt]
\multicolumn{6}{@{}l}{\textbf{Qwen2.5 1.5B}} \\
Base model & \multicolumn{3}{c}{one fixed run} & \underline{55.33} &  \\
NTP post-training & 54.86 & 54.98 & 54.98 & 54.94 & $\pm$0.17 \\
Rand. synonyms & 54.64 & 54.57 & 54.64 & 54.61 & $\pm$0.09 \\
Conceptizer & 55.48 & 55.37 & 55.35 & \textbf{55.40} & $\pm$0.17 \\
\addlinespace[3pt]
\multicolumn{6}{@{}l}{\textbf{Qwen2.5 3B}} \\
Base model & \multicolumn{3}{c}{one fixed run} & \underline{55.87} &  \\
NTP post-training & 55.43 & 55.47 & 55.51 & 55.47 & $\pm$0.10 \\
Rand. synonyms & 54.69 & 55.00 & 54.86 & 54.85 & $\pm$0.38 \\
Conceptizer & 56.10 & 56.00 & 56.26 & \textbf{56.12} & $\pm$0.32 \\
\addlinespace[3pt]
\multicolumn{6}{@{}l}{\textbf{Qwen2.5 7B}} \\
Base model & \multicolumn{3}{c}{one fixed run} & 55.94 &  \\
NTP post-training & 56.15 & 56.25 & 56.22 & \underline{56.21} & $\pm$0.12 \\
Rand. synonyms & 55.52 & 55.53 & 55.47 & 55.51 & $\pm$0.08 \\
Conceptizer & 56.82 & 56.72 & 56.75 & \textbf{56.76} & $\pm$0.12 \\
\addlinespace[3pt]
\multicolumn{6}{@{}l}{\textbf{Falcon3 1B}} \\
Base model & \multicolumn{3}{c}{one fixed run} & \underline{57.12} &  \\
NTP post-training & 56.61 & 56.64 & 56.67 & 56.64 & $\pm$0.07 \\
Rand. synonyms & 56.74 & 56.51 & 56.63 & 56.63 & $\pm$0.29 \\
Conceptizer & 57.54 & 57.59 & 57.48 & \textbf{57.54} & $\pm$0.13 \\
\addlinespace[3pt]
\multicolumn{6}{@{}l}{\textbf{Falcon3 3B}} \\
Base model & \multicolumn{3}{c}{one fixed run} & \textbf{58.14} &  \\
NTP post-training & 57.06 & 57.07 & 57.04 & 57.06 & $\pm$0.03 \\
Rand. synonyms & 56.41 & 56.39 & 56.57 & 56.46 & $\pm$0.25 \\
Conceptizer & 57.95 & 57.76 & 57.86 & \underline{57.86} & $\pm$0.24 \\
\addlinespace[3pt]
\multicolumn{6}{@{}l}{\textbf{Falcon3 7B}} \\
Base model & \multicolumn{3}{c}{one fixed run} & \underline{57.58} &  \\
NTP post-training & 57.40 & 57.48 & 57.48 & 57.45 & $\pm$0.11 \\
Rand. synonyms & 56.49 & 56.54 & 56.54 & 56.52 & $\pm$0.08 \\
Conceptizer & 58.17 & 58.31 & 58.31 & \textbf{58.26} & $\pm$0.20 \\
\bottomrule
\end{tabular}
\caption{\textbf{Sentiment group at every training seed: Conceptizer has the best mean against the 3 baselines in 8 of 9 models.} The mean of IMDb, ToxicConversations, TweetSentimentExtraction, Emotion, ToxicChat, PoemSentiment, FrenkEn, YelpReviewFull and AmazonPolarity; 95\% CI: margin of error of the Student-$t$ training-seed interval. Rand. synonyms: Randomized synonyms.}\label{tab:seed-sentiment}
\end{table}

\begin{table}[tp]\centering\footnotesize\setlength{\tabcolsep}{3pt}\renewcommand{\arraystretch}{1}
\begin{tabular}[t]{@{}lrrrrr@{}}
\toprule
 & \multicolumn{5}{c}{Accuracy (\%) $\uparrow$} \\
\cmidrule(l){2-6}
Method & Seed 0 & 37 & 42 & Mean & 95\% CI \\
\midrule
\multicolumn{6}{@{}l}{\textbf{Llama 1B}} \\
Base model & \multicolumn{3}{c}{one fixed run} & 67.02 &  \\
NTP post-training & 66.72 & 67.15 & 66.87 & 66.91 & $\pm$0.54 \\
Rand. synonyms & 67.80 & 67.43 & 67.39 & \underline{67.54} & $\pm$0.55 \\
Conceptizer & 68.71 & 68.18 & 69.20 & \textbf{68.69} & $\pm$1.27 \\
\addlinespace[3pt]
\multicolumn{6}{@{}l}{\textbf{Llama 3B}} \\
Base model & \multicolumn{3}{c}{one fixed run} & 74.16 &  \\
NTP post-training & 75.81 & 75.30 & 75.09 & \underline{75.40} & $\pm$0.92 \\
Rand. synonyms & 75.32 & 74.23 & 74.89 & 74.81 & $\pm$1.36 \\
Conceptizer & 75.78 & 75.53 & 75.16 & \textbf{75.49} & $\pm$0.78 \\
\addlinespace[3pt]
\multicolumn{6}{@{}l}{\textbf{Llama 8B}} \\
Base model & \multicolumn{3}{c}{one fixed run} & \textbf{82.41} &  \\
NTP post-training & 82.04 & 82.25 & 81.81 & \underline{82.03} & $\pm$0.55 \\
Rand. synonyms & 81.15 & 80.57 & 81.23 & 80.98 & $\pm$0.89 \\
Conceptizer & 82.28 & 81.05 & 81.61 & 81.64 & $\pm$1.53 \\
\addlinespace[3pt]
\multicolumn{6}{@{}l}{\textbf{Qwen2.5 1.5B}} \\
Base model & \multicolumn{3}{c}{one fixed run} & 71.11 &  \\
NTP post-training & 72.10 & 72.60 & 73.04 & \textbf{72.58} & $\pm$1.17 \\
Rand. synonyms & 71.18 & 71.49 & 71.50 & 71.39 & $\pm$0.46 \\
Conceptizer & 72.31 & 72.40 & 72.14 & \underline{72.28} & $\pm$0.32 \\
\addlinespace[3pt]
\multicolumn{6}{@{}l}{\textbf{Qwen2.5 3B}} \\
Base model & \multicolumn{3}{c}{one fixed run} & 76.44 &  \\
NTP post-training & 77.10 & 77.47 & 76.67 & \textbf{77.08} & $\pm$1.00 \\
Rand. synonyms & 75.35 & 74.92 & 74.56 & 74.94 & $\pm$0.98 \\
Conceptizer & 76.14 & 76.97 & 76.99 & \underline{76.70} & $\pm$1.21 \\
\addlinespace[3pt]
\multicolumn{6}{@{}l}{\textbf{Qwen2.5 7B}} \\
Base model & \multicolumn{3}{c}{one fixed run} & 80.13 &  \\
NTP post-training & 81.89 & 82.21 & 81.79 & \textbf{81.97} & $\pm$0.55 \\
Rand. synonyms & 80.62 & 80.71 & 80.46 & 80.60 & $\pm$0.32 \\
Conceptizer & 81.73 & 82.05 & 81.85 & \underline{81.88} & $\pm$0.40 \\
\addlinespace[3pt]
\multicolumn{6}{@{}l}{\textbf{Falcon3 1B}} \\
Base model & \multicolumn{3}{c}{one fixed run} & 65.79 &  \\
NTP post-training & 65.94 & 65.84 & 65.95 & \underline{65.91} & $\pm$0.15 \\
Rand. synonyms & 65.19 & 65.23 & 65.09 & 65.17 & $\pm$0.18 \\
Conceptizer & 66.08 & 66.30 & 66.46 & \textbf{66.28} & $\pm$0.47 \\
\addlinespace[3pt]
\multicolumn{6}{@{}l}{\textbf{Falcon3 3B}} \\
Base model & \multicolumn{3}{c}{one fixed run} & \textbf{69.90} &  \\
NTP post-training & 69.03 & 69.32 & 69.19 & \underline{69.18} & $\pm$0.36 \\
Rand. synonyms & 68.08 & 67.98 & 67.31 & 67.79 & $\pm$1.04 \\
Conceptizer & 68.51 & 67.88 & 67.91 & 68.10 & $\pm$0.88 \\
\addlinespace[3pt]
\multicolumn{6}{@{}l}{\textbf{Falcon3 7B}} \\
Base model & \multicolumn{3}{c}{one fixed run} & \textbf{78.56} &  \\
NTP post-training & 78.04 & 78.52 & 78.44 & \underline{78.33} & $\pm$0.64 \\
Rand. synonyms & 77.07 & 76.71 & 76.81 & 76.86 & $\pm$0.46 \\
Conceptizer & 77.75 & 77.71 & 77.61 & 77.69 & $\pm$0.17 \\
\bottomrule
\end{tabular}
\caption{\textbf{Commonsense group at every training seed: the best mean belongs to Conceptizer, NTP post-training and the base model in 3 models each.} The mean of HellaSwag, COPA and WinoGrande; 95\% CI: margin of error of the Student-$t$ training-seed interval. Rand. synonyms: Randomized synonyms.}\label{tab:seed-commonsense}
\end{table}

\begin{table}[tp]\centering\footnotesize\setlength{\tabcolsep}{3pt}\renewcommand{\arraystretch}{1}
\begin{tabular}[t]{@{}lcccc@{}}
\toprule
 & \multicolumn{4}{c}{Knowledge accuracy (\%) $\uparrow$} \\
\cmidrule(l){2-5}
Method & ARC-C & ARC-E & OBQA & SciQ \\
\midrule
\multicolumn{5}{@{}l}{\textbf{Llama 1B}} \\
Base model & 37.71 & 63.64 & 37.60 & 92.20 \\
NTP post-training & 37.91 & 64.91 & 38.13 & 91.70 \\
Randomized synonyms & \underline{38.40} & \underline{65.52} & \underline{39.07} & \underline{92.23} \\
Conceptizer & \textbf{38.68} & \textbf{66.02} & \textbf{40.13} & \textbf{92.67} \\
\addlinespace[3pt]
\multicolumn{5}{@{}l}{\textbf{Llama 3B}} \\
Base model & \underline{45.99} & 71.38 & 39.20 & 93.20 \\
NTP post-training & 45.36 & \underline{71.80} & 40.87 & \textbf{94.40} \\
Randomized synonyms & \textbf{46.05} & 70.93 & \underline{41.47} & 94.20 \\
Conceptizer & 45.96 & \textbf{72.31} & \textbf{43.20} & \underline{94.33} \\
\addlinespace[3pt]
\multicolumn{5}{@{}l}{\textbf{Llama 8B}} \\
Base model & \textbf{55.29} & \textbf{79.76} & \underline{45.00} & \underline{95.80} \\
NTP post-training & 52.84 & \underline{79.17} & 44.47 & 95.63 \\
Randomized synonyms & 53.67 & 78.10 & 43.87 & 95.57 \\
Conceptizer & \underline{54.38} & 78.55 & \textbf{45.07} & \textbf{96.27} \\
\addlinespace[3pt]
\multicolumn{5}{@{}l}{\textbf{Qwen2.5 1.5B}} \\
Base model & 46.50 & \underline{75.84} & 40.20 & \textbf{94.10} \\
NTP post-training & \underline{46.64} & 75.36 & 40.73 & 93.10 \\
Randomized synonyms & 46.22 & 75.35 & \underline{42.13} & \underline{93.43} \\
Conceptizer & \textbf{48.27} & \textbf{76.84} & \textbf{42.53} & 93.17 \\
\addlinespace[3pt]
\multicolumn{5}{@{}l}{\textbf{Qwen2.5 3B}} \\
Base model & 48.04 & 72.81 & 41.60 & 91.30 \\
NTP post-training & 49.15 & 76.50 & 42.27 & 95.27 \\
Randomized synonyms & \underline{49.69} & \underline{76.81} & \underline{45.00} & \underline{95.30} \\
Conceptizer & \textbf{51.22} & \textbf{78.75} & \textbf{45.47} & \textbf{95.33} \\
\addlinespace[3pt]
\multicolumn{5}{@{}l}{\textbf{Qwen2.5 7B}} \\
Base model & 55.29 & 80.93 & 48.80 & 93.60 \\
NTP post-training & 57.25 & \underline{82.15} & 48.67 & 95.03 \\
Randomized synonyms & \underline{57.37} & 81.97 & \underline{49.20} & \textbf{95.37} \\
Conceptizer & \textbf{58.67} & \textbf{82.95} & \textbf{49.33} & \underline{95.07} \\
\addlinespace[3pt]
\multicolumn{5}{@{}l}{\textbf{Falcon3 1B}} \\
Base model & \underline{45.31} & 68.01 & 40.60 & 93.10 \\
NTP post-training & 44.65 & 69.77 & 40.60 & 94.07 \\
Randomized synonyms & 44.48 & \underline{70.83} & \underline{41.53} & \textbf{94.33} \\
Conceptizer & \textbf{46.79} & \textbf{72.00} & \textbf{43.20} & \underline{94.27} \\
\addlinespace[3pt]
\multicolumn{5}{@{}l}{\textbf{Falcon3 3B}} \\
Base model & \textbf{54.27} & 74.87 & 42.40 & \textbf{95.40} \\
NTP post-training & 49.72 & \textbf{76.23} & 41.87 & \underline{95.13} \\
Randomized synonyms & 50.63 & 74.45 & \underline{42.47} & 94.47 \\
Conceptizer & \underline{52.47} & \underline{75.90} & \textbf{43.87} & 94.97 \\
\addlinespace[3pt]
\multicolumn{5}{@{}l}{\textbf{Falcon3 7B}} \\
Base model & \underline{59.73} & 77.86 & 45.60 & \textbf{94.80} \\
NTP post-training & 58.48 & \textbf{79.59} & 44.47 & \underline{94.67} \\
Randomized synonyms & 57.88 & 78.40 & \underline{45.73} & 94.57 \\
Conceptizer & \textbf{59.95} & \underline{79.41} & \textbf{48.07} & 94.13 \\
\bottomrule
\end{tabular}
\caption{\textbf{Conceptizer has the best knowledge accuracy on OpenBookQA in all 9 models and on ARC-Challenge and ARC-Easy in 6 each.} Means of training seeds 0, 37 and 42.}\label{tab:tasks-knowledge}
\end{table}

\begin{table}[tp]\centering\footnotesize\setlength{\tabcolsep}{3pt}\renewcommand{\arraystretch}{1}
\begin{tabular}[t]{@{}lccc@{}}
\toprule
 & \multicolumn{3}{c}{Commonsense accuracy (\%) $\uparrow$} \\
\cmidrule(l){2-4}
Method & HellaSwag & COPA & WinoGrande \\
\midrule
\multicolumn{4}{@{}l}{\textbf{Llama 1B}} \\
Base model & 61.57 & 78.00 & 61.48 \\
NTP post-training & \textbf{61.96} & 77.00 & \underline{61.77} \\
Randomized synonyms & 61.12 & \underline{80.33} & 61.17 \\
Conceptizer & \underline{61.74} & \textbf{82.33} & \textbf{62.01} \\
\addlinespace[3pt]
\multicolumn{4}{@{}l}{\textbf{Llama 3B}} \\
Base model & 71.65 & 82.00 & 68.82 \\
NTP post-training & \textbf{71.93} & \underline{84.33} & \textbf{69.93} \\
Randomized synonyms & 71.55 & \underline{84.33} & 68.56 \\
Conceptizer & \underline{71.85} & \textbf{85.67} & \underline{68.96} \\
\addlinespace[3pt]
\multicolumn{4}{@{}l}{\textbf{Llama 8B}} \\
Base model & \textbf{79.59} & \textbf{94.00} & 73.64 \\
NTP post-training & \underline{79.28} & \underline{93.00} & \underline{73.82} \\
Randomized synonyms & 78.53 & 91.33 & 73.09 \\
Conceptizer & 78.75 & 92.33 & \textbf{73.85} \\
\addlinespace[3pt]
\multicolumn{4}{@{}l}{\textbf{Qwen2.5 1.5B}} \\
Base model & \textbf{68.26} & 82.00 & 63.06 \\
NTP post-training & \underline{67.39} & \textbf{86.00} & 64.35 \\
Randomized synonyms & 66.99 & 82.67 & \underline{64.51} \\
Conceptizer & 67.38 & \underline{84.33} & \textbf{65.14} \\
\addlinespace[3pt]
\multicolumn{4}{@{}l}{\textbf{Qwen2.5 3B}} \\
Base model & \textbf{74.96} & 85.00 & 69.38 \\
NTP post-training & \underline{73.61} & \textbf{87.33} & \underline{70.30} \\
Randomized synonyms & 72.87 & 81.67 & \underline{70.30} \\
Conceptizer & 73.45 & \underline{85.67} & \textbf{70.98} \\
\addlinespace[3pt]
\multicolumn{4}{@{}l}{\textbf{Qwen2.5 7B}} \\
Base model & \textbf{80.44} & 89.00 & 70.96 \\
NTP post-training & \underline{79.08} & \underline{92.33} & \textbf{74.48} \\
Randomized synonyms & 78.18 & 90.00 & 73.61 \\
Conceptizer & 78.80 & \textbf{92.67} & \underline{74.16} \\
\addlinespace[3pt]
\multicolumn{4}{@{}l}{\textbf{Falcon3 1B}} \\
Base model & \textbf{63.14} & 74.00 & \underline{60.22} \\
NTP post-training & 62.38 & \underline{75.00} & \textbf{60.35} \\
Randomized synonyms & 61.57 & \underline{75.00} & 58.93 \\
Conceptizer & \underline{62.66} & \textbf{76.67} & 59.51 \\
\addlinespace[3pt]
\multicolumn{4}{@{}l}{\textbf{Falcon3 3B}} \\
Base model & \textbf{70.15} & \textbf{75.00} & \underline{64.56} \\
NTP post-training & \underline{67.89} & \underline{74.67} & \textbf{64.98} \\
Randomized synonyms & 66.35 & 74.33 & 62.69 \\
Conceptizer & 67.18 & 73.67 & 63.46 \\
\addlinespace[3pt]
\multicolumn{4}{@{}l}{\textbf{Falcon3 7B}} \\
Base model & \textbf{78.43} & \textbf{87.00} & 70.24 \\
NTP post-training & \underline{76.48} & \underline{86.67} & \textbf{71.85} \\
Randomized synonyms & 75.20 & 85.33 & 70.06 \\
Conceptizer & 75.54 & 86.00 & \underline{71.53} \\
\bottomrule
\end{tabular}
\caption{\textbf{The base model has the best HellaSwag accuracy in 7 of 9 models, Conceptizer the best COPA accuracy in 4, and NTP post-training the best WinoGrande accuracy in 5 (Conceptizer in the other 4).} Means of training seeds 0, 37 and 42.}\label{tab:tasks-commonsense}
\end{table}

\begin{table*}[tp]\centering\footnotesize\setlength{\tabcolsep}{3pt}\renewcommand{\arraystretch}{1}
\begin{tabular}[t]{@{}lccccccccc@{}}
\toprule
 & \multicolumn{9}{c}{Test-split accuracy (\%) $\uparrow$} \\
\cmidrule(l){2-10}
Method & IMDb & ToxConv. & TweetSent. & Emotion & ToxicChat & Poem & FrenkEn & Yelp & Amazon \\
\midrule
\multicolumn{10}{@{}l}{\textbf{Llama 1B}} \\
Base model & \textbf{68.40} & 65.47 & \underline{51.88} & \textbf{33.28} & \underline{67.63} & 42.31 & \underline{58.72} & \textbf{51.76} & \underline{72.52} \\
NTP post-training & 67.27 & 65.16 & 51.86 & 32.14 & 66.38 & \underline{42.44} & 58.67 & \underline{51.30} & 71.59 \\
Randomized synonyms & 66.36 & \underline{65.92} & 51.82 & 32.17 & 66.04 & 41.79 & 58.19 & 51.26 & 70.53 \\
Conceptizer & \underline{68.04} & \textbf{66.38} & \textbf{52.22} & \underline{33.21} & \textbf{68.57} & \textbf{43.33} & \textbf{59.09} & 51.21 & \textbf{72.74} \\
\addlinespace[3pt]
\multicolumn{10}{@{}l}{\textbf{Llama 3B}} \\
Base model & \underline{69.48} & 65.21 & \underline{51.95} & \textbf{32.86} & \underline{67.68} & 40.67 & 59.25 & \textbf{53.32} & \underline{72.25} \\
NTP post-training & 68.89 & 65.07 & 51.93 & 31.57 & 67.36 & \underline{40.96} & \underline{59.30} & \underline{53.01} & 72.12 \\
Randomized synonyms & 67.37 & \underline{65.51} & 51.79 & 31.75 & 67.50 & 40.83 & 59.07 & 52.94 & 70.28 \\
Conceptizer & \textbf{70.01} & \textbf{66.05} & \textbf{52.28} & \underline{32.51} & \textbf{69.63} & \textbf{42.85} & \textbf{60.14} & 52.44 & \textbf{73.36} \\
\addlinespace[3pt]
\multicolumn{10}{@{}l}{\textbf{Llama 8B}} \\
Base model & \underline{70.24} & 63.67 & \underline{52.68} & \underline{31.53} & \underline{68.10} & 38.65 & 59.21 & \textbf{54.30} & \underline{74.15} \\
NTP post-training & 69.85 & \underline{64.53} & 52.58 & 31.51 & 66.63 & \underline{39.04} & \underline{60.11} & 53.97 & 74.06 \\
Randomized synonyms & 67.18 & 64.34 & 52.19 & 31.29 & 66.77 & 38.78 & 59.73 & \underline{54.19} & 72.02 \\
Conceptizer & \textbf{71.08} & \textbf{64.68} & \textbf{52.79} & \textbf{32.10} & \textbf{68.65} & \textbf{40.64} & \textbf{60.23} & 54.09 & \textbf{75.59} \\
\addlinespace[3pt]
\multicolumn{10}{@{}l}{\textbf{Qwen2.5 1.5B}} \\
Base model & \textbf{68.61} & 62.08 & 51.57 & 29.76 & \textbf{64.50} & 43.08 & \underline{55.77} & \underline{52.35} & \textbf{70.23} \\
NTP post-training & 67.72 & \underline{62.18} & \underline{51.73} & \underline{30.20} & 62.97 & 42.88 & \textbf{55.81} & 51.66 & 69.31 \\
Randomized synonyms & 66.65 & 61.86 & 51.26 & 29.66 & 62.65 & \underline{43.53} & 55.41 & \textbf{52.37} & 68.14 \\
Conceptizer & \underline{68.03} & \textbf{62.78} & \textbf{51.95} & \textbf{30.72} & \underline{63.64} & \textbf{44.01} & 55.68 & 51.79 & \underline{69.99} \\
\addlinespace[3pt]
\multicolumn{10}{@{}l}{\textbf{Qwen2.5 3B}} \\
Base model & \underline{69.42} & \underline{64.16} & \underline{52.08} & \underline{29.86} & 64.93 & \textbf{41.35} & 56.75 & 52.41 & \textbf{71.85} \\
NTP post-training & 68.54 & 63.95 & 51.90 & 29.40 & \underline{64.99} & 40.06 & \underline{56.80} & 52.57 & 71.00 \\
Randomized synonyms & 67.01 & 63.89 & 51.25 & 29.57 & 64.94 & 38.01 & 56.77 & \textbf{53.16} & 69.02 \\
Conceptizer & \textbf{69.55} & \textbf{64.81} & \textbf{52.17} & \textbf{30.19} & \textbf{66.19} & \underline{40.61} & \textbf{57.42} & \underline{52.66} & \underline{71.46} \\
\addlinespace[3pt]
\multicolumn{10}{@{}l}{\textbf{Qwen2.5 7B}} \\
Base model & \textbf{71.03} & \textbf{63.88} & 50.81 & 29.28 & \underline{68.15} & 38.08 & 57.57 & 52.17 & \underline{72.48} \\
NTP post-training & 70.04 & 62.81 & \underline{50.96} & \underline{29.82} & 66.96 & \underline{42.15} & \underline{58.14} & \underline{52.62} & 72.36 \\
Randomized synonyms & 66.93 & 63.17 & 50.47 & 29.77 & 67.30 & 41.09 & 57.41 & \textbf{53.24} & 70.20 \\
Conceptizer & \underline{70.22} & \underline{63.56} & \textbf{51.03} & \textbf{30.56} & \textbf{68.28} & \textbf{43.40} & \textbf{58.31} & 52.39 & \textbf{73.09} \\
\addlinespace[3pt]
\multicolumn{10}{@{}l}{\textbf{Falcon3 1B}} \\
Base model & \underline{70.12} & 66.01 & 51.79 & 31.08 & 69.64 & \textbf{42.98} & \textbf{57.63} & \underline{52.00} & \underline{72.81} \\
NTP post-training & 69.10 & 65.84 & \underline{52.54} & \underline{31.44} & 68.91 & 41.89 & 56.19 & 51.44 & 72.41 \\
Randomized synonyms & 68.30 & \underline{66.09} & 52.15 & 31.31 & \underline{69.81} & \underline{42.69} & 56.08 & \textbf{52.13} & 71.08 \\
Conceptizer & \textbf{70.46} & \textbf{66.34} & \textbf{52.73} & \textbf{32.59} & \textbf{70.58} & 42.63 & \underline{56.94} & 51.51 & \textbf{74.04} \\
\addlinespace[3pt]
\multicolumn{10}{@{}l}{\textbf{Falcon3 3B}} \\
Base model & \textbf{71.25} & \textbf{65.61} & 51.81 & 31.74 & \textbf{73.45} & \textbf{43.17} & \textbf{59.24} & 51.90 & \underline{75.13} \\
NTP post-training & \underline{69.98} & 64.77 & \underline{52.78} & \underline{31.82} & 68.02 & 41.79 & 58.41 & 51.64 & 74.32 \\
Randomized synonyms & 68.23 & 64.40 & 52.24 & 31.81 & 68.41 & 41.19 & 57.59 & \textbf{52.40} & 71.85 \\
Conceptizer & \textbf{71.25} & \underline{64.88} & \textbf{53.03} & \textbf{33.13} & \underline{69.46} & \underline{42.88} & \underline{58.67} & \underline{51.94} & \textbf{75.48} \\
\addlinespace[3pt]
\multicolumn{10}{@{}l}{\textbf{Falcon3 7B}} \\
Base model & \textbf{72.95} & \underline{65.92} & 51.75 & 30.57 & \textbf{68.78} & 40.38 & \underline{59.19} & \textbf{52.96} & \underline{75.70} \\
NTP post-training & 71.94 & 65.05 & \underline{52.33} & \underline{30.77} & 67.54 & \underline{43.04} & 58.72 & 52.30 & 75.39 \\
Randomized synonyms & 68.57 & 65.34 & 51.67 & 30.46 & 66.84 & 42.05 & 58.77 & 52.47 & 72.54 \\
Conceptizer & \underline{72.77} & \textbf{65.97} & \textbf{52.87} & \textbf{31.77} & \underline{67.92} & \textbf{44.68} & \textbf{59.33} & \underline{52.53} & \textbf{76.53} \\
\bottomrule
\end{tabular}
\caption{\textbf{Conceptizer has the best sentiment accuracy on TweetSentimentExtraction in all 9 models, on the 7 other tasks in 5 to 7 each, and on YelpReviewFull in none.} Means of training seeds 0, 37 and 42.}\label{tab:tasks-sentiment}
\end{table*}

\begin{table*}[tp]\centering\footnotesize\setlength{\tabcolsep}{3pt}\renewcommand{\arraystretch}{1}
\begin{tabular}[t]{@{}lrrrrrrrrr@{}}
\toprule
 & \multicolumn{9}{c}{Held-out perplexity $\downarrow$} \\
\cmidrule(l){2-10}
 & \multicolumn{3}{c}{Concept-target} & \multicolumn{3}{c}{Content-word} & \multicolumn{3}{c}{All tokens} \\
\cmidrule(lr){2-4}\cmidrule(lr){5-7}\cmidrule(l){8-10}
Method & Seed 0 & 37 & 42 & Seed 0 & 37 & 42 & Seed 0 & 37 & 42 \\
\midrule
\multicolumn{10}{@{}l}{\textbf{Llama 1B}} \\
Base model & \multicolumn{3}{c}{1229.58} & \multicolumn{3}{c}{500.78} & \multicolumn{3}{c}{21.12} \\
NTP post-training & 1014.70 & 1012.41 & 1018.06 & 384.95 & 384.51 & 384.94 & 17.84 & 17.85 & 17.85 \\
Rand. synonyms & 548.33 & 551.42 & 551.19 & 331.06 & 331.49 & 331.61 & 19.88 & 19.88 & 19.88 \\
Conceptizer & 367.16 & 370.37 & 369.72 & 373.71 & 373.73 & 374.16 & 19.31 & 19.32 & 19.31 \\
\addlinespace[3pt]
\multicolumn{10}{@{}l}{\textbf{Llama 3B}} \\
Base model & \multicolumn{3}{c}{760.82} & \multicolumn{3}{c}{297.02} & \multicolumn{3}{c}{17.08} \\
NTP post-training & 616.36 & 617.81 & 617.13 & 225.41 & 225.85 & 225.45 & 14.47 & 14.47 & 14.47 \\
Rand. synonyms & 339.40 & 339.76 & 339.68 & 203.98 & 203.80 & 203.84 & 16.18 & 16.18 & 16.18 \\
Conceptizer & 218.14 & 218.27 & 218.44 & 225.52 & 225.84 & 225.40 & 15.68 & 15.68 & 15.68 \\
\addlinespace[3pt]
\multicolumn{10}{@{}l}{\textbf{Llama 8B}} \\
Base model & \multicolumn{3}{c}{391.41} & \multicolumn{3}{c}{145.11} & \multicolumn{3}{c}{13.11} \\
NTP post-training & 338.68 & 340.66 & 338.46 & 118.15 & 117.97 & 118.11 & 11.39 & 11.39 & 11.39 \\
Rand. synonyms & 199.04 & 200.17 & 198.09 & 111.67 & 111.45 & 111.51 & 12.62 & 12.62 & 12.62 \\
Conceptizer & 128.97 & 128.85 & 128.12 & 119.56 & 119.67 & 119.40 & 12.19 & 12.20 & 12.20 \\
\addlinespace[3pt]
\multicolumn{10}{@{}l}{\textbf{Qwen2.5 1.5B}} \\
Base model & \multicolumn{3}{c}{723.06} & \multicolumn{3}{c}{275.97} & \multicolumn{3}{c}{14.28} \\
NTP post-training & 663.10 & 664.37 & 663.30 & 247.50 & 247.80 & 247.76 & 13.73 & 13.73 & 13.73 \\
Rand. synonyms & 361.35 & 362.86 & 362.12 & 206.21 & 206.14 & 206.01 & 15.07 & 15.07 & 15.07 \\
Conceptizer & 231.71 & 231.22 & 230.95 & 232.09 & 232.25 & 232.02 & 14.48 & 14.48 & 14.48 \\
\addlinespace[3pt]
\multicolumn{10}{@{}l}{\textbf{Qwen2.5 3B}} \\
Base model & \multicolumn{3}{c}{551.47} & \multicolumn{3}{c}{198.26} & \multicolumn{3}{c}{12.69} \\
NTP post-training & 478.69 & 480.38 & 479.81 & 174.88 & 174.91 & 174.84 & 12.16 & 12.16 & 12.16 \\
Rand. synonyms & 269.38 & 267.79 & 268.43 & 150.56 & 150.58 & 150.69 & 13.34 & 13.33 & 13.33 \\
Conceptizer & 167.77 & 167.39 & 167.27 & 165.61 & 165.48 & 165.61 & 12.81 & 12.80 & 12.81 \\
\addlinespace[3pt]
\multicolumn{10}{@{}l}{\textbf{Qwen2.5 7B}} \\
Base model & \multicolumn{3}{c}{435.07} & \multicolumn{3}{c}{145.12} & \multicolumn{3}{c}{11.33} \\
NTP post-training & 358.80 & 359.11 & 359.39 & 124.32 & 124.35 & 124.38 & 10.77 & 10.77 & 10.77 \\
Rand. synonyms & 199.65 & 200.63 & 200.50 & 112.08 & 111.98 & 112.20 & 11.77 & 11.77 & 11.78 \\
Conceptizer & 128.09 & 128.63 & 128.22 & 120.86 & 120.59 & 120.52 & 11.34 & 11.34 & 11.34 \\
\addlinespace[3pt]
\multicolumn{10}{@{}l}{\textbf{Falcon3 1B}} \\
Base model & \multicolumn{3}{c}{1377.40} & \multicolumn{3}{c}{651.94} & \multicolumn{3}{c}{18.94} \\
NTP post-training & 1127.16 & 1128.16 & 1127.95 & 501.12 & 500.89 & 500.74 & 16.04 & 16.04 & 16.04 \\
Rand. synonyms & 639.96 & 639.27 & 639.39 & 393.09 & 393.26 & 393.34 & 18.21 & 18.21 & 18.21 \\
Conceptizer & 306.40 & 307.15 & 306.83 & 410.92 & 410.82 & 410.99 & 17.38 & 17.38 & 17.37 \\
\addlinespace[3pt]
\multicolumn{10}{@{}l}{\textbf{Falcon3 3B}} \\
Base model & \multicolumn{3}{c}{1259.24} & \multicolumn{3}{c}{554.76} & \multicolumn{3}{c}{17.59} \\
NTP post-training & 917.72 & 918.48 & 917.56 & 389.43 & 389.54 & 389.61 & 14.80 & 14.80 & 14.80 \\
Rand. synonyms & 509.78 & 509.63 & 510.11 & 318.23 & 318.24 & 318.25 & 16.79 & 16.79 & 16.79 \\
Conceptizer & 242.86 & 243.42 & 242.87 & 327.66 & 327.45 & 327.40 & 16.06 & 16.06 & 16.06 \\
\addlinespace[3pt]
\multicolumn{10}{@{}l}{\textbf{Falcon3 7B}} \\
Base model & \multicolumn{3}{c}{980.95} & \multicolumn{3}{c}{314.38} & \multicolumn{3}{c}{15.09} \\
NTP post-training & 555.47 & 554.70 & 555.21 & 193.90 & 193.86 & 193.97 & 12.03 & 12.03 & 12.03 \\
Rand. synonyms & 334.06 & 333.99 & 334.84 & 175.48 & 175.68 & 175.89 & 13.67 & 13.67 & 13.67 \\
Conceptizer & 151.04 & 151.28 & 150.84 & 170.36 & 170.56 & 170.37 & 13.05 & 13.06 & 13.06 \\
\bottomrule
\end{tabular}
\caption{\textbf{Held-out perplexity at every training seed: Conceptizer has the lowest perplexity against the 3 baselines at concept-target positions, and NTP post-training the lowest over all tokens.} 3-seed means in Fig.~\ref{fig:ppl-matrix}. Rand. synonyms: Randomized synonyms.}\label{tab:seed-perplexity}
\end{table*}

\clearpage
\SizeAccuracyFigure
\TrainingCurvesFigure

\end{document}